\documentclass[sigconf]{acmart}
\usepackage{subfig}
\usepackage{multirow}
\usepackage{balance}

\copyrightyear{2025}
\acmYear{2025}
\setcopyright{acmlicensed}
\acmConference[MM '25] {Proceedings of the 33rd ACM International Conference on Multimedia}{October 27--31, 2025}{Dublin, Ireland.}
\acmBooktitle{Proceedings of the 33rd ACM International Conference on Multimedia (MM '25), October 27--31, 2025, Dublin, Ireland}
\acmISBN{979-8-4007-2035-2/2025/10}
\acmDOI{10.1145/3746027.3755261}

\begin{document}

\title{BridgeNet: A Unified Multimodal Framework for Bridging 2D and 3D Industrial Anomaly Detection}


\author{An Xiang}
\authornote{Equally contribute to this work.}
\affiliation{%
  \institution{ Shenzhen Institutes of Advanced Technology, Chinese Academy of Sciences}
  \institution{ University of Chinese Academy of Sciences}
  \city{Shenzhen}
  \country{China}
}
\email{a.xiang@siat.ac.cn}

\author{Zixuan Huang}
\authornotemark[1]
\affiliation{%
  \institution{ Shenzhen Institutes of Advanced Technology, Chinese Academy of Sciences}
  \institution{ University of Chinese Academy of Sciences}
  \city{Shenzhen}
  \country{China}
}
\email{zx.huang5@siat.ac.cn}

\author{Xitong Gao}
\authornotemark[1]
\affiliation{%
  \institution{ Shenzhen Institutes of Advanced Technology, Chinese Academy of Sciences}
  \institution{Shenzhen University of Advanced Technology}
  \city{Shenzhen}
  \country{China}
}
\email{xt.gao@siat.ac.cn}

\author{Kejiang Ye}
\authornote{Corresponding author.}
\affiliation{
  \institution{Shenzhen Institutes of Advanced Technology, Chinese Academy of Sciences}
  \city{Shenzhen}
  \country{China}
}
\email{kj.ye@siat.ac.cn}

\author{Cheng-zhong Xu}
\affiliation{%
  \institution{ State Key Lab of IOTSC, Department of CIS, University of Macau}
  \city{Macau SAR}
  \country{China}
}
\email{czxu@um.edu.mo}

%
\renewcommand{\shortauthors}{An Xiang, Zixuan Huang, Xitong Gao, Kejiang Ye, Cheng-zhong Xu}

\begin{abstract}
Industrial anomaly detection for 2D objects has gained significant attention and achieved progress in anomaly detection (AD) methods. However, identifying 3D depth anomalies using only 2D information is insufficient. Despite explicitly fusing depth information into RGB images or using point cloud backbone networks to extract depth features, both approaches struggle to adequately represent 3D information in multimodal scenarios due to the disparities among different modal information. Additionally, due to the scarcity of abnormal samples in industrial data, especially in multimodal scenarios, it is necessary to perform anomaly generation to simulate real-world abnormal samples. Therefore, we propose a novel unified multimodal anomaly detection framework to address these issues. Our contributions consist of 3 key aspects. 
(1) We extract visible depth information from 3D point cloud data simply and use 2D RGB images to represent appearance, which disentangles depth and appearance to support unified anomaly generation. 
(2) Benefiting from the flexible input representation, the proposed Multi-Scale Gaussian Anomaly Generator and Unified Texture Anomaly Generator can generate richer anomalies in RGB and depth. 
(3) All modules share parameters for both RGB and depth data, effectively bridging 2D and 3D anomaly detection. 
Subsequent modules can directly leverage features from both modalities without complex fusion. Experiments show our method outperforms state-of-the-art (SOTA) on MVTec-3D AD and Eyecandies datasets. Code available at: \url{https://github.com/Xantastic/BridgeNet}

\end{abstract}




\maketitle

\section{Introduction}
\label{sec:intro}
Industrial anomaly detection is the basis of quality control, which has been increasingly explored with the development of deep learning technologies. 
However, industrial anomaly detection encounters two key challenges: (1) the scarcity of anomalous samples, and (2) the diverse types of anomalies, ranging from slight changes to significant structural anomalies~\cite{liu1,bergmann2019mvtec,zou2022spot}.
\begin{figure}[t!]
    \centering
    \includegraphics[width=1.00\linewidth]{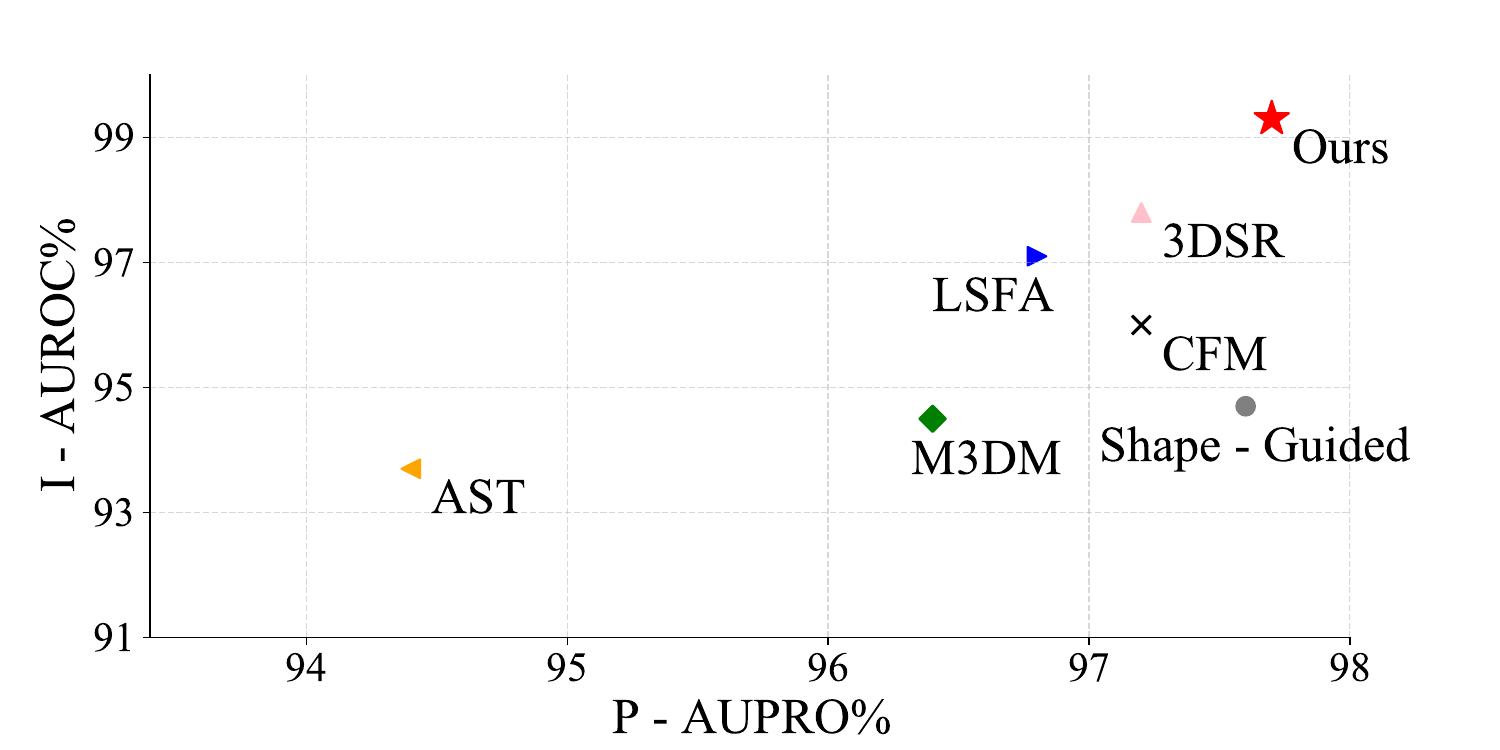}
    \caption{I-AUROC and P-AUPRO on MVTec-3D AD benchmark: Our BridgeNet outperforms all previous methods in detection and localization.}
    \label{fig:enter-label}
\end{figure}

Given the two key challenges in industrial anomaly detection, supervised methods present practical limitations, prompting current research to focus on unsupervised approaches. Unsupervised anomaly detection methods based on 2D data are categorized into two types according to their reliance on data augmentation:

\textbf{(1) Without augmentation}: Embedding-based methods use ImageNet pre-trained models for feature extraction and statistical approaches to model normal product distributions~\cite{roth2, rudolph3}.Then, anomalies are detected by comparing new input features with the modeled distribution via nearest neighbor~\cite{cohen4, defard5, liu2023diversity}. Reconstruction-based methods train only on normal samples, assuming anomalies are poorly reconstructed during testing, enabling detection and localization via pixel-wise reconstruction errors~\cite{perera9, zavrtanik6, gong7, bergmann8}. Knowledge distillation-based methods have student networks learn only teacher networks' normal-data outputs. Anomalies are then estimated by comparing their output differences~\cite{deng10,tien11}.

\textbf{(2) With augmentation}: Recent works have utilized various augmentation strategies to introduce synthetic anomalies. These approaches seek to expand the diversity of anomalous features to enhance the model's capacity to distinguish anomalies from normal data distributions~\cite{zavrtanik12, li13, zavrtanik2022dsr}. Although generating-based methods~\cite{chen16} have demonstrated SOTA performance, they applied to RGB images~\cite{liu1, roth2, batzner14, zhou15, chen16} still perform poorly in detecting depth anomalies. The reason is while color information aids in identifying most anomalies, accurately detecting depth anomalies requires 3D geometric information~\cite{horwitz18}.


Although recent works on 3D anomaly generation \cite{Li33, zavrtanik2024keep} have been done, they solely generate local anomalies using Perlin noise at the raw data or feature level. These synthetic 3D anomalies exhibit significant disparities from real-world anomalies both in terms of data appearance and feature representation. More importantly, real anomalous data manifests distinguishable differences from normal data after undergoing data preprocessing and feature extraction. However, existing approaches \cite{{zavrtanik30, Li33, zavrtanik2024keep}} typically simulate these differences by directly injecting noise into the model at a single fixed location, which may fail to adequately capture the diverse distribution patterns of real anomalies.


To address the previously mentioned challenges, we propose a novel unified framework for anomaly detection and localization, called BridgeNet. First, instead of directly concatenating the 3D information to the 2D pre-trained features, we transform the point cloud into a depth image and then utilize a 2D pre-trained model to extract the 3D features. Then we feed the concatenation of both modalities features to the fusion adaptor. This approach brings the representation distributions of both modalities closer and transforms their information within a shared target domain. Subsequently, we propose two dual-modal anomaly generation methods. The Multi-scale Gaussian Anomaly Generator adds different scales Gaussian noise at multiple locations in the model, following the adding large-to-small noise for shallow-to-deep model levels. This generator is more in line with the scale variations of data in the model, and the generated anomaly semantics are richer. Unified Texture Anomaly Generator adds texture anomalies to both RGB and depth images, bridging 2D and 3D anomalies through textures. Benefiting from the diversity of the Describable Textures Dataset (DTD) \cite{cimpoi2014describing}, the anomalies we generate are richer and more realistic \cite{Li33, zavrtanik2024keep}. Finally, for both RGB and depth images, the parameters of all the modules in model are shared. In addition, existing 3D feature extraction methods can not efficiently align with RGB images \cite{cao34, chu21, wang20, costanzino35}. The parameter-sharing framework implicitly aligns both modalities without additional modal alignment, and the feature distributions are very close and similar, as shown in Figure ~\ref{fig8}.  
The main contributions of this work are summarized below:
\begin{figure}[t!]
    \centering
    \subfloat[Rope Distribution]{\includegraphics[scale=0.12]{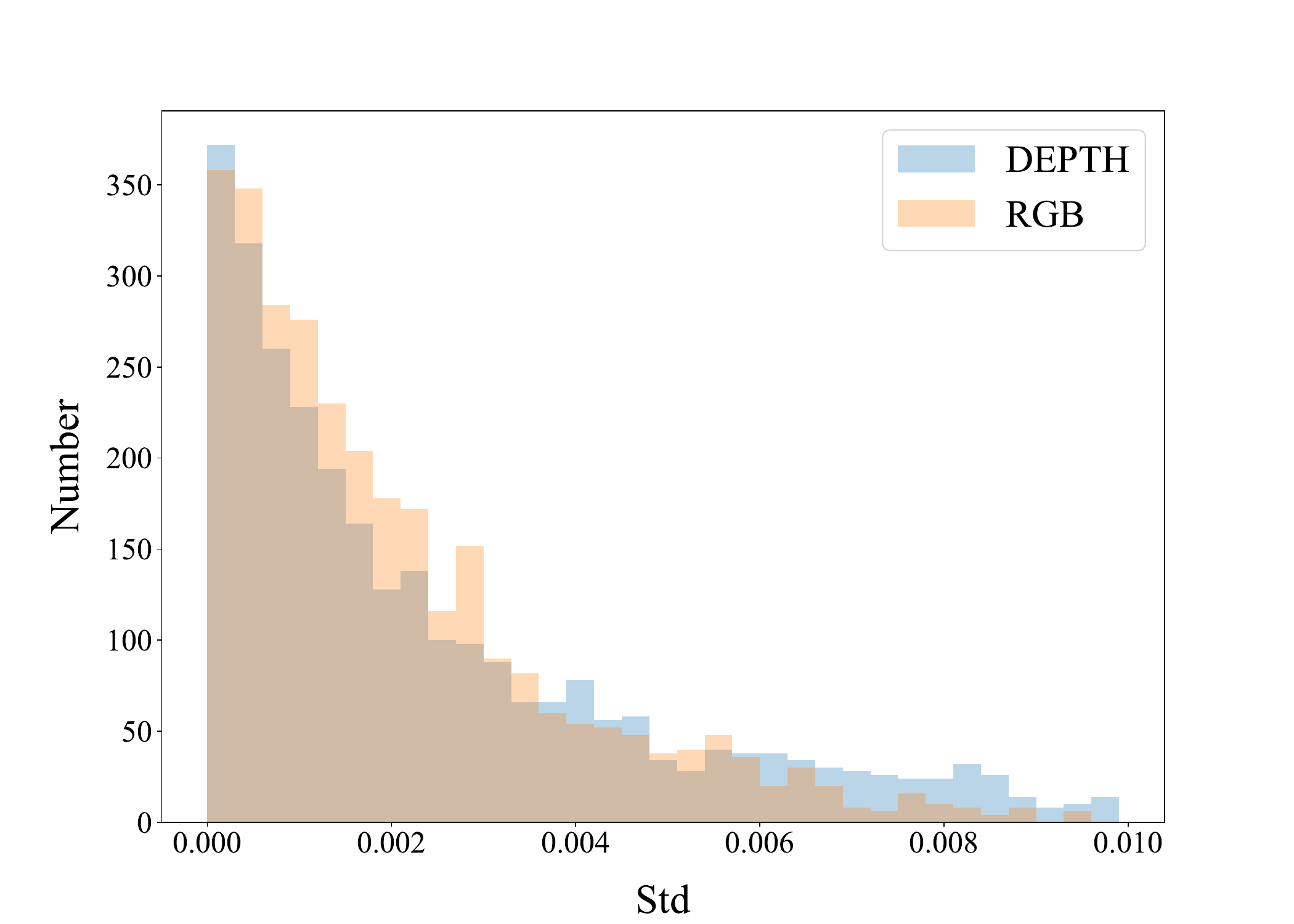}}
    \subfloat[Foam Distribution]{\includegraphics[scale=0.12]{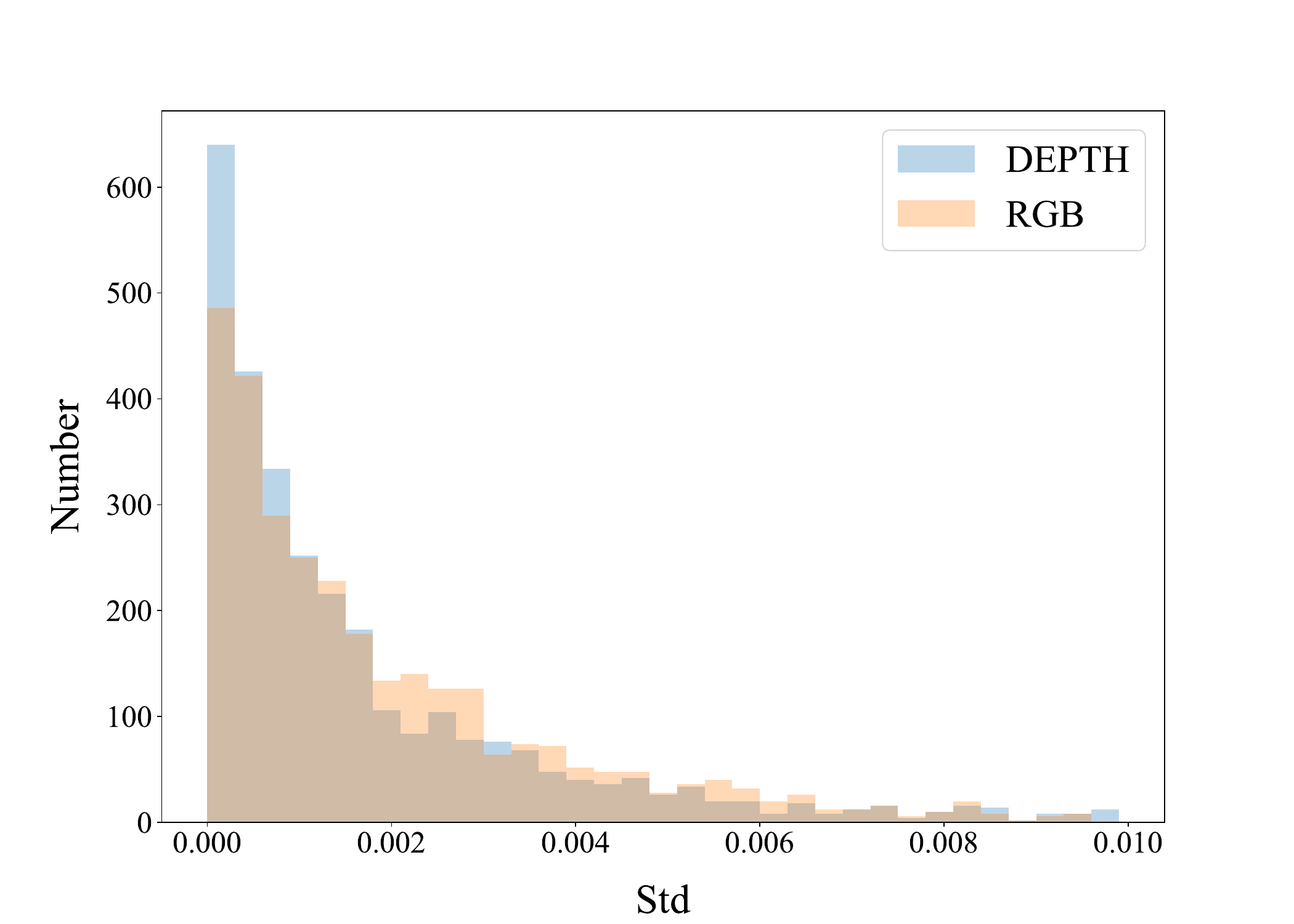}}
    \caption{Distribution of features extracted in 3D and 2D information through the shared-parameter Feature Extractor.}
    \label{fig8} 
\end{figure}
\begin{itemize}
\item A \textbf{Unified Parameter-sharing Framework for RGB and depth images.} The framework implicitly aligns both modalities, achieving close feature distribution alignment without additional alignment modules.
\item A \textbf{Multi-scale Gaussian Anomaly Generator (MGAG)} that applies multi-scale Gaussian noise at multiple model levels, adding large-to-small noise for shallow-to-deep model levels. This design better captures the scale variations of data within the model and enriches the semantic diversity of generated anomalies.
\item A \textbf{Unified Texture Anomaly Generator (UTAG)} that applies texture anomalies from DTD \cite{cimpoi2014describing} to both RGB and depth images, focusing on appearance and effectively bridging 2D and 3D anomalies.
\item Our proposed framework achieves SOTA performance on the MVTec-3D AD dataset, as shown in Figure ~\ref{fig:enter-label}, with I-AUROC of 99.3\% and P-AUPRO of 97.7\%.
\end{itemize}

\section{Related Work}
\label{sec:work}
\subsection{2D Anomaly Detection}
Embedding-based Methods feed anomaly-free images into ImageNet-based pre-trained models to extract features, which are then embedded into an normal feature distribution using statistical algorithms during the training phase~\cite{cohen4, defard5, roth2}. 
These methods detect anomalies by comparing the input features with the learned anomaly-free distribution and using kNN for anomaly score computation during the testing phase.
The primary difference from the reconstruction-based methods is that they detect anomalies in a high-dimensional feature space rather than in the RGB image space.
Flow-based Methods use normalizing flow to transform the distribution of anomaly-free samples into a simple distribution (\textit{e.g.} Gaussian distribution)~\cite{serra23, rudolph3, lei2023pyramidflow}. 
Anomalous samples that have not appeared in training are transformed away from the learned distribution and assigned a high anomaly score.
Distillation-based Methods train student models using only anomaly-free samples to match the output of frozen weight pre-trained teacher models~\cite{bergmann27, deng10, gu2023remembering, wu2024aekd}. 
Since student models are not trained on anomalous images, they usually cannot mimic teacher models on these images.

Although many improved versions have also appeared, subtle depth anomalies are difficult to detect using 2D modality methods in industrial scenarios~\cite{horwitz18, bergmann17}.
Therefore, researchers have begun to explore strategies for 3D.

\subsection{3D Anomaly Detection}
While 3D anomaly detection faces greater technical challenges compared to 2D approaches, it holds considerable promise. Proper utilization of depth data can significantly improve detection performance in targeted applications.
With numerous methods proposed for only RGB-based anomaly detection, recent researches are now working to bridge these concepts to 3D anomaly detection.

Horwitz~\textit{et al.}~\cite{horwitz18} noted rotation-invariant 3D representations are crucial for 3D anomaly detection, proposing the BTF method that combines color and geometric attributes for better 3D dataset performance.
Rudolph~\textit{et al.}~\cite{rudolph19} proposed the S-T model for 3D datasets, with flow-based teacher and CNN-based student networks. However, it directly concatenates 3D and RGB data without 3D feature extraction, leading to poor depth representation.
3DSR~\cite{zavrtanik30}, the current SOTA, is a 3D surface anomaly detection approach based on dual-subspace reprojection but only uses single-scale Perlin noise to simulate anomalies, often insufficient in multimodal scenarios.
CFM~\cite{costanzino35} detects anomalies in multimodal data through learning cross-modal feature mappings between 2D and 3D modalities.
LSFA and M3DM~\cite{wang20, tu2024self} applied two backbone networks, pre-trained on RGB and point cloud, and fused the extracted features to store them into multiple memory banks. However, Point cloud data is difficult to align with RGB data, and using a dual-branch network incredases computational complexity.
Although EasyNet~\cite{chen32} transforms point-cloud information into depth images for input, it uses independent auto-encoders to reconstruct different modal information. Despite a fusion module for integration, interference from inconsistent modal information distribution causes low detection accuracy.

BridgeNet addresses the above multimodal anomaly detection challenges. It introduces a unified parameter-sharing framework to implicitly align RGB and depth images without additional modules, and applies multi-scale noise within the model to detect distribution deviations at different locations (before and after fusion). Furthermore, converting anomalous point cloud data into depth maps reveals distinct texture anomalies; thus, BridgeNet extends 2D texture anomaly generation methods to 3D depth images to simulate more realistic depth anomalies.

\begin{figure*}[h]
    \centering
    \includegraphics[width=0.9\linewidth]{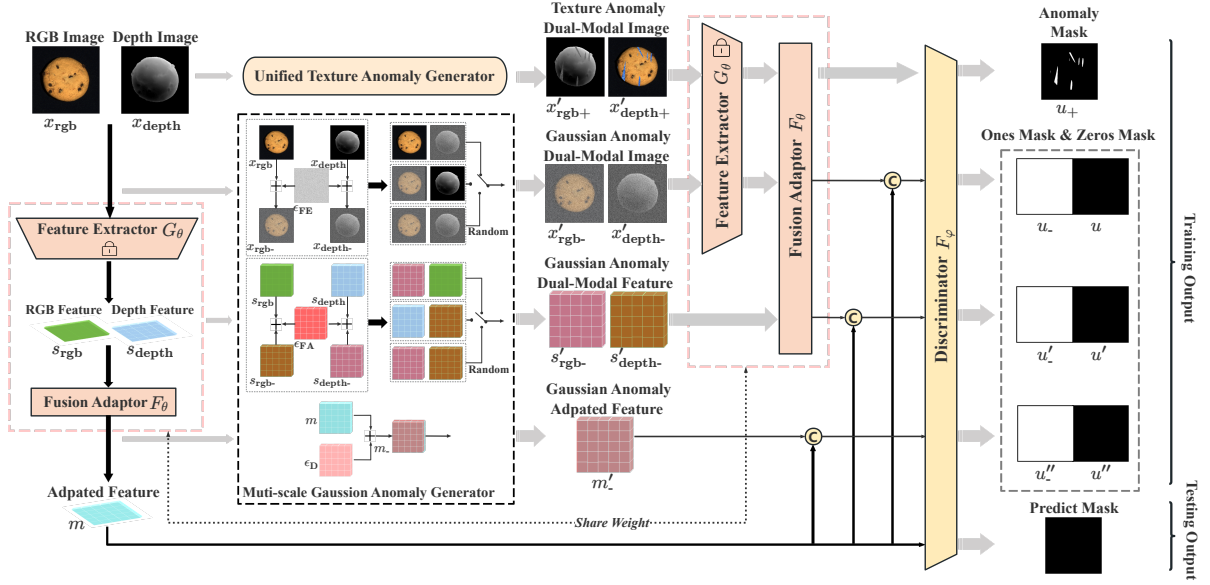}
    \caption{Overview of the proposed BridgeNet. In the training phase, RGB and depth images are inputted into shared \textit{Feature Extractor} to obtain their respective features. These features are then concatenated to feed the \textit{Fusion Adaptor}. The \textit{MGAG} generates Gaussian anomalies, and the \textit{UTAG} introduces texture anomalies, each producing dual-modal anomalous features. Adapted anomaly features and their mask are used to train \textit{Dual-modal Discriminator}. During inference, the anomaly generators are removed, and the trained Discriminator outputs predictions based on the adapted features alone.}
    \label{fig3}
\end{figure*}
\section{Method}
As depicted in Figure ~\ref{fig3}, BridgeNet consists of a \textit{Multi-scale Gaussian Anomaly Generator (MGAG)}, a \textit{Unified Texture Anomaly Generator (UTAG)} and a \textit{Dual-modal Discriminator}. 
The parameters of all the modules in model are shared for both modalities. 
The following sections will describe these modules.
\subsection{Feature Extractor and Fusion Adaptor}
The features of RGB and depth images are extracted using a ResNet-like model pre-trained on ImageNet, which has proven effective~\cite{liu1, roth2, chen16}.
We proposed using a parameter-sharing pre-trained model for both modalities, which implicitly aligns both modalities without additional modal alignment. The feature distributions of both modalities are similar and close.
For any RGB image $x_{\textrm{rgb}, i} \in \mathbb{R}^{H \times W \times 3}$ in $\mathcal{X}_{\textrm{rgb}, train}$ and depth image $x_{\textrm{depth}, i} \in \mathbb{R}^{H \times W \times 3}$ in $\mathcal{X}_{\textrm{depth}, train}$, the pre-trained network $\phi$ extracts the $j$th layer features as $v_{\textrm{rgb}, j, i}=\phi_{j}\left(x_{\textrm{rgb}, i}\right), v_{\textrm{depth}, j, i}=\phi_{j}\left(x_{\textrm{depth}, i}\right)$. 


We denote the RGB and depth patch features at the $j$th layer's $(h, w)$ position as $p_{\textrm{rgb}, j, i}^{h, w}$ and $p_{\textrm{depth}, j, i}^{h, w}$, derived from the $(h, w)$ features $v_{\textrm{rgb}, j, i}^{h, w}$, $v_{\textrm{depth}, j, i}^{h, w}$ and their neighborhood features via adaptive average pooling. Their neighborhood is expressed as:
\begin{equation}
\begin{aligned}
    \mathcal{N}_{\textrm{M}, p}^{(h, w)}=\{&(a, b) \mid a \in[h-\lfloor p / 2\rfloor, \ldots, h+\lfloor p / 2\rfloor],\\
    & b \in[w-\lfloor p / 2\rfloor, \ldots, w+\lfloor p / 2\rfloor]\},
\end{aligned}
\end{equation}
where $p$ is patch size and $\textrm{M} \in {\textrm{rgb},\textrm{depth}}$. Then, the aggregated neighborhood features by aggregation function can be given by the following formulation:
\begin{equation}
\begin{aligned}
    p_{\textrm{M}, j, i}^{h, w}=f_{\textrm{agg}}\left(\left\{v_{\textrm{M}, j, i}^{h^{\prime}, w^{\prime}} \mid\left(h^{\prime}, w^{\prime}\right) \in \mathcal{N}_{\textrm{M},p}^{(h, w)}\right\}\right),
\end{aligned}
\end{equation}
where $M \in {\textrm{rgb},\textrm{depth}}$. We then upsample the depth features to size $(h_0, w_0)$ as the shallowest feature by nearest interpolation and concatenate to get the final feature map $s_{\textrm{rgb}, i}, s_{\textrm{depth}, i} \in \mathbb{R}^{H_{0} \times W_{0} \times C}$.
\begin{equation}
\begin{aligned}
    s_{\textrm{M}, i}=f_{\textrm{cat}}\left(\operatorname{resize}\left(p_{\textrm{M}, j^{\prime}, i}\left(H_{0}, W_{0}\right)\right), j^{\prime} \in J\right.,
\end{aligned}
\end{equation}
where $M \in {\textrm{rgb},\textrm{depth}}$. All the above processes are denoted as $G_\theta$.
Benefiting from shared weights, RGB and depth are implicitly aligned without additional process.
We directly concatenate $s_{\textrm{rgb}, i} $ and $s_{\textrm{depth}, i}$ on the feature dimension to obtain $o_i$.
Then, we use a Fusion Adaptor $F_{\theta}$, consisting of a simple fully connected layer, to fuse and transfer data from two different modalities to the target domain, where each fused feature is represented as $d_{i}=F_{\theta}\left(o_{i}\right)$.
\subsection{Multi-Scale Gaussian Anomaly Generator}
Gaussian noise has been added to generate anomalies in various ways. However, most current generated Gaussian anomalies are only performed at a single fixed model layer, which may cause the model to pay insufficient attention to anomalies perturbations at other semantic levels. We propose to generate anomalies at different semantic abstraction levels by adding multi-scale of Gaussian noise to the original image and before and after the Fusion Adaptor. This helps the model learn richer feature representations and enhances recognition and understanding of anomalies. Our experiments show that adding large-to-small anomalies for shallow-to-deep level features yields better performance results. Based on this above assumption, we design MGAG as follows.

\textbf{Multi-scale Gaussian noise} During training, we define our designed Gaussian noise for Feature Extractor, Fusion Adaptor and Discriminator as:
\begin{equation}
\begin{aligned}
    \epsilon_{\textrm{FE},\textrm{M},i} \sim \mathcal{N}\left(u_{1}, \sigma_{1}\right), \quad
    \epsilon_{\textrm{FA},\textrm{M},i} \sim \mathcal{N}\left(u_{2}, \sigma_{2}\right), \quad
    \epsilon_{\textrm{D},i} \sim \mathcal{N}\left(u_{3}, \sigma_{3}\right),
\end{aligned}   
\end{equation}
where $\sigma_{1}>\sigma_{2}>\sigma_{3}, u_{1}=u_{2}=u_{3}=0$ and $\textrm{M} \in {\textrm{rgb},\textrm{depth}}$. Then, we send the noise to the inputs of Feature Extractor $x_{\textrm{rgb}, i}, x_{\textrm{depth}, i}$, Fusion Adaptor $s_{\textrm{rgb}, i}, s_{\textrm{depth}, i}$ and Discriminator $d_i$. The multi-scale Gaussian anomalies are defined as follows:
\begin{equation}
\begin{aligned}
    x_{\textrm{M}, i-}=x_{\textrm{M}, i}+\epsilon_{\textrm{FE}, \textrm{M}, i},\quad 
    s_{\textrm{M}, i-}=s_{\textrm{M}, i}+\epsilon_{\textrm{FA}, \textrm{M}, i},\quad 
    d_{i-}=d_{i}+\epsilon_{\textrm{D}, i},
\end{aligned}
\end{equation}
where $i-$ means the Gaussian anomalies of $i$th product. We add Gaussian noise to $x$, $s$ and $d$ in each model level to generate multi-scale Gaussian anomalies.

\textbf{Selective Modality} In real-world scenarios, ensuring both modalities capture anomalies is hard. Considering only dual-modality anomalies may make the model over-rely on their significant features, reducing generalization and robustness. Inspired by GLASS \cite{chen16} randomly selecting three anomaly region combinations to enrich texture anomaly diversity, we randomly select one of three modality anomaly combinations per epoch to enhance dual-modal anomaly diversity. Thus, we propose Selective Modality:

\begin{equation}
    \begin{aligned}
        x_{\textrm{rgb}, i-}^{\prime}, x_{\textrm{depth}, i-}^{\prime} &= 
        \begin{cases}
            x_{\textrm{rgb}, i-}, x_{\textrm{depth}, i-}, & \text{if} \quad 1 > p > 2\alpha, \\
            x_{\textrm{rgb}, i-}, x_{\textrm{depth}, i},    & \text{if} \quad 2\alpha > p > \alpha, \\
            x_{\textrm{rgb}, i}, x_{\textrm{depth}, i-},    & \text{if} \quad \alpha > p > 0,
        \end{cases} \\
        d_{i-}^{\prime} &= d_{i-},
    \end{aligned}
\end{equation}
where $\alpha$ is set to 1/3, and $x_{\textrm{rgb}, i-}^{\prime}$, $x_{\textrm{depth}, i-}^{\prime}$ and $d_{i-}^{\prime}$ are final generated Gaussian anomalies. The expression for $s_{\textrm{rgb}, i-}^{\prime}, s_{\textrm{depth}, i-}^{\prime}$ follows the same pattern as $x$.

\begin{table*}[!t]
    \centering
    \caption{BridgeNet achieves superior I-AUROC scores for anomaly detection across all MVTec-3D AD categories, outperforming SOTA works in 3D, RGB, and combined settings.}
    \scalebox{1.0}{\begin{tabular}{@{}llccccccccccc@{}}
        \toprule
          & \textbf{Method} & \textbf{Bagel} & \textbf{Cable Gland} & \textbf{Carrot} & \textbf{Cookie} & \textbf{Dowel} & \textbf{Foam} & \textbf{Peach} & \textbf{Potato} & \textbf{Rope} & \textbf{Tire} & \textbf{Mean} \\
        \midrule
        \multirow{6}*{\rotatebox[origin=c]{90}{\textbf{3D}}} & EasyNet \cite{chen32} & 0.735 & 0.678 & 0.747 & 0.864 & 0.719 & 0.716 & 0.713 & 0.725 & 0.885 & 0.687 & 0.747 \\
        ~ & AST \cite{rudolph19} & 0.881 & 0.576 & 0.965 & 0.957 & 0.679 & 0.797 & 0.990 & 0.915 & 0.956 & 0.611 & 0.833 \\
        ~ & M3DM \cite{wang20} & 0.941 & 0.651 & 0.965 & 0.969 & 0.905 & 0.760 & 0.880 & 0.974 & 0.926 & 0.765 & 0.874 \\
        ~ & Shape-Guided \cite{chu21} & 0.983 & 0.682 & 0.978 & \textbf{0.998} & \textbf{0.960} & 0.737 & 0.993 & \textbf{0.979} & 0.966 & 0.871 & 0.916 \\
        ~ & 3DSR \cite{zavrtanik30} & 0.945 & \textbf{0.835} & 0.969 & 0.857 & 0.955 & 0.880 & 0.963 & 0.934 & \textbf{0.998} & \textbf{0.888} & 0.922 \\
        ~ & LSFA \cite{tu2024self} & \textbf{0.986}& 0.669& 0.973& 0.990& 0.950& 0.802& 0.961& 0.964& 0.967& 0.944& 0.921\\
        ~ & BridgeNet (Ours) & 0.983 & 0.817 & \textbf{0.984} & 0.981 & 0.909 & \textbf{0.903} & \textbf{0.994} & 0.951 & 0.986 & 0.838 & \textbf{0.935} \\
        \midrule
        \multirow{6}*{\rotatebox[origin=c]{90}{\textbf{RGB}}} & EasyNet \cite{chen32} & \textbf{0.982} & 0.992 & 0.917 & \textbf{0.953} & 0.919 & 0.923 & 0.840 & 0.785 & 0.986 & 0.742 & 0.904 \\
        ~ & AST \cite{rudolph19} & 0.947 & 0.928 & 0.851 & 0.825 & 0.981 & \textbf{0.951} & 0.895 & 0.613 & \textbf{0.992} & 0.821 & 0.880 \\
        ~ & M3DM \cite{wang20} & 0.944 & 0.918 & 0.896 & 0.749 & 0.959 & 0.767 & 0.919 & 0.648 & 0.938 & 0.767 & 0.850 \\
        ~ & Shape-Guided \cite{chu21} & 0.911 & 0.936 & 0.883 & 0.662 & 0.974 & 0.772 & 0.785 & 0.641 & 0.884 & 0.706 & 0.815 \\
        ~ & 3DSR \cite{zavrtanik30} & 0.844 & 0.930 & 0.964 & 0.794 & \textbf{0.998} & 0.904 & 0.938 & 0.730 & 0.978 & 0.900 & 0.898 \\
        ~ & LSFA \cite{tu2024self} & 0.951& 0.920& 0.911& 0.762& 0.961& 0.770& 0.930& 0.675& 0.938& 0.787& 0.861\\
        ~ & BridgeNet (Ours) & 0.981 & \textbf{0.960} & \textbf{0.979} & 0.801 & 0.997 & 0.851 & \textbf{0.941} & \textbf{0.864} & 0.956 & \textbf{0.916} & \textbf{0.925} \\
        \midrule
        \multirow{6}*{\rotatebox[origin=c]{90}{\textbf{3D+RGB}}} & EasyNet \cite{chen32} & 0.991 & \textbf{0.998} & 0.918 & 0.968 & 0.945 & 0.945 & 0.905 & 0.807 & 0.994 & 0.793 & 0.926 \\
        ~ & AST \cite{rudolph19} & 0.983 & 0.873 & 0.976 & 0.971 & 0.932 & 0.885 & 0.974 & 0.981 & 1.000 & 0.797 & 0.937 \\
        ~ & M3DM \cite{wang20} & 0.994 & 0.909 & 0.972 & 0.976 & 0.960 & 0.942 & 0.973 & 0.899 & 0.972 & 0.850 & 0.945 \\
        ~ & Shape-Guided \cite{chu21} & 0.986 & 0.894 & 0.983 & 0.991 & 0.976 & 0.857 & 0.990 & 0.965 & 0.960 & 0.869 & 0.947 \\
        ~ & 3DSR \cite{zavrtanik30} & 0.981 & 0.867 & 0.996 & 0.981 & \textbf{1.000} & \textbf{0.994} & 0.986 & 0.978 & \textbf{1.000} & \textbf{0.995} & 0.978 \\
        ~ & CFM \cite{costanzino35} & 0.988& 0.875& 0.984& 0.992& 0.997& 0.924& 0.964& 0.949& 0.979& 0.950& 0.960\\
        ~ & LSFA \cite{tu2024self} & \textbf{1.000} & 0.939& 0.982& 0.989& 0.961& 0.951& 0.983& 0.962& 0.989& 0.951& 0.971\\
        ~ & BridgeNet (Ours) & 0.999 & \textbf{0.998} & \textbf{1.000} & \textbf{0.995} & 0.999 & 0.991 & \textbf{0.995} & \textbf{0.998} & 0.998 & 0.961 & \textbf{0.993} \\
        \bottomrule
    \end{tabular}}
    \label{tab1}
\end{table*}

\subsection{Unified Texture Anomaly Generator}
\begin{figure}[t]
    \centering
    \includegraphics[width=0.9\linewidth]{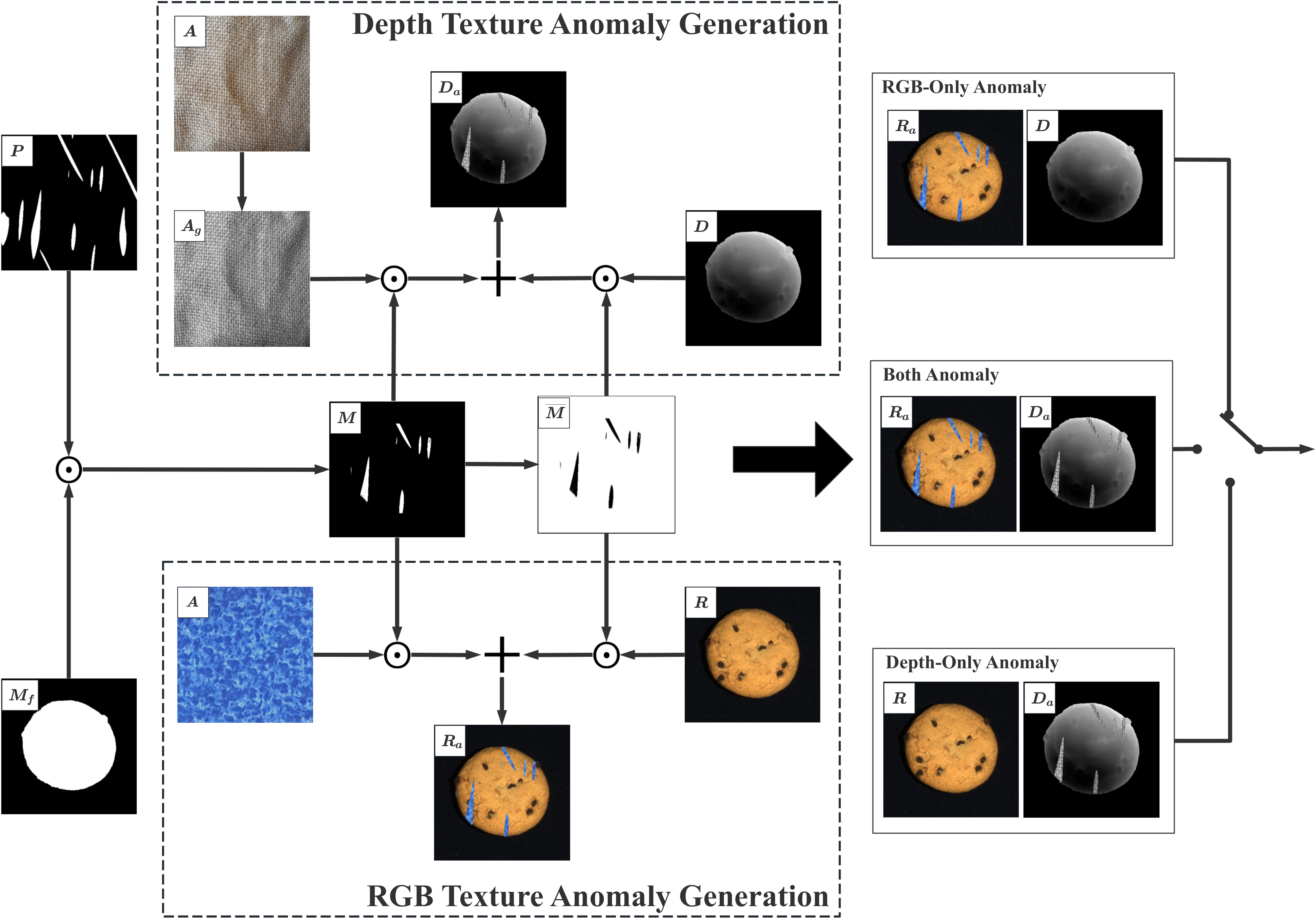}
    \caption{Diagram of the Unified Texture Anomaly Generator.}
    \label{fig5}
\end{figure}

In the texture anomaly generation stage, anomalies diffuse to various feature representation positions as the model propagation goes deeper. 
If noises are added at the feature level, it is unpredictable which positions in the representation of texture anomalous images will undergo significant changes. 
Thus, only image-level anomaly generation is used for texture anomaly generation.
Inspired by the 2D anomaly generation method \cite{chen16, zavrtanik30}, we provide a unified anomaly generation framework, bridging 2D and 3D anomalies by texture. The detailed flow is shown in Figure ~\ref{fig5}.

\textbf{RGB image texture anomaly generation} We refer to the effective texture feature generation method \cite{chen16, zavrtanik30}.
The texture anomaly mask $m_l$ is derived by the intersection operation between the generated Perlin noise-based mask $m_p$ and the foreground mask~$m_f$. 
The anomaly patch $t_{\textrm{rgb}, i}$ uses a randomly selected DTD dataset \cite{cimpoi2014describing} image, enriched via augmentation.
RGB image texture anomaly $x_{\textrm{rgb}, i+}$ can be defined as:


\begin{equation}
    x_{\textrm{rgb}, i+}=x_{\textrm{rgb}, i} \odot \bar{m}_{t}+(1-\beta) t_{\textrm{rgb}, i} \odot m_{t}+\beta x_{\textrm{rgb}, i} \odot m_{t},
\end{equation}
where $\bar{m}_{t}$ is the pixel-wise inverse operation of ${m}_{t}$, $\odot$ is Hadamard product and $\beta$ is an opacity parameter representing different degrees of anomalies.

\textbf{Depth image texture anomaly generation} As the point cloud information is converted to depth images, the depth anomalys (\textit{e.g.}, holes) are all represented as visible texture anomaly on the three-channel images, as shown in Figure ~\ref{fig5}. Thus, We generate texture anomalies on depth images similar to RGB ones. Furthermore, since the three-channel depth image has the same value for each channel, it behaves as grayscale. Thus, for the anomalous patches from the DTD dataset \cite{cimpoi2014describing}, first convert them to grayscale $x_{\textrm{depth}, i}$ and then feed the depth anomaly images into the same feature space as RGB images through a shared-parameter network. The subsequent process is similar to RGB image texture anomaly generation. The depth image texture anomaly as:
\begin{equation}
    x_{\textrm{depth}, i+}=x_{\textrm{depth}, i} \odot \bar{m}_{t}+(1-\beta) t_{\textrm{depth}, i} \odot m_{t}+\beta x_{\textrm{depth}, i} \odot m_{t}.
\end{equation}

Building on the Selective Modality concept, real-world industrial anomalies like RGB color variations or depth holes often cannot be adequately represented in the opposite modality. Thus, we generate texture anomalies using a random combination of three anomaly types. The texture anomalies $x_{\textrm{rgb}, i+}, x_{\textrm{depth}, i+}$ are defined as follows:
\begin{equation}
    x_{\textrm{rgb}, i+}^{\prime} , x_{\textrm{depth}, i+}^{\prime} = \begin{cases}
    x_{\textrm{rgb}, i+}, x_{\textrm{depth}, i+} & \text{if} \quad 1>p>2 \alpha, \\
    x_{\textrm{rgb}, i+}, x_{\textrm{depth}, i} & \text{if} \quad 2 \alpha>p>\alpha, \\
    x_{\textrm{rgb}, i}, x_{\textrm{depth}, i+} & \text{if} \quad \alpha>p>0.
\end{cases}
\end{equation}

\begin{table*}[h]
    \centering
    \caption{BridgeNet achieves superior P-AUPRO scores for anomaly localization across all MVTec-3D AD categories, outperforming SOTA works in 3D, RGB, and combined settings.}
    \scalebox{1.0}{\begin{tabular}{@{}llccccccccccc@{}}
        \toprule
          & \textbf{Method} & \textbf{Bagel} & \textbf{Cable Gland} & \textbf{Carrot} & \textbf{Cookie} & \textbf{Dowel} & \textbf{Foam} & \textbf{Peach} & \textbf{Potato} & \textbf{Rope} & \textbf{Tire} & \textbf{Mean} \\
        \midrule
        \multirow{5}*{\rotatebox[origin=c]{90}{\textbf{3D}}} & EasyNet \cite{chen32} & 0.160 & 0.030 & 0.680 & 0.759 & 0.758 & 0.069 & 0.225 & 0.734 & 0.797 & 0.509 & 0.472 \\
        ~ & M3DM \cite{wang20} & 0.943 & 0.818 & 0.977 & 0.882 & 0.881 & 0.743 & 0.958 & 0.974 & 0.950 & 0.929 & 0.906 \\
        ~ & Shape-Guided \cite{chu21} & 0.974 & 0.871 & 0.981 & 0.924 & 0.898 & 0.773 & 0.978 & \textbf{0.983} & 0.955 & \textbf{0.969} & 0.931 \\
        ~ & 3DSR \cite{zavrtanik30} & 0.922 & 0.872 & \textbf{0.984} & 0.859 & 0.940 & 0.714 & 0.970 & 0.978 &  \textbf{0.977} & 0.858 & 0.907 \\
        ~ & LSFA \cite{tu2024self} & 0.974& 0.887& 0.981& 0.921& 0.901& 0.773& 0.982& 0.983& 0.959& 0.981& 0.934\\
        ~ & BridgeNet (Ours) & \textbf{0.978} & \textbf{0.931} & 0.983 & \textbf{0.963} & \textbf{0.939} & \textbf{0.798} & \textbf{0.983} & \textbf{0.983} & 0.973 & 0.923 & \textbf{0.945} \\
        \midrule
        \multirow{5}*{\rotatebox[origin=c]{90}{\textbf{RGB}}} & EasyNet \cite{chen32} & 0.751 & 0.825 & 0.916 & 0.599 & 0.698 & 0.699 & 0.917 & 0.827 & 0.887 & 0.636 & 0.776 \\
        ~ & M3DM \cite{wang20} & 0.952 & 0.972 & 0.973 & 0.891 & 0.932 & 0.843 & 0.970 & 0.956 & 0.968 & 0.966 & 0.942 \\
        ~ & Shape-Guided \cite{chu21} & 0.946 & 0.972 & 0.960 & 0.914 & 0.958 & 0.776 & 0.937 & 0.949 & 0.956 & 0.957 & 0.933 \\
        ~ & 3DSR \cite{zavrtanik30} & 0.923 & 0.970 & 0.979 & 0.859 & \textbf{0.979} & \textbf{0.894} & 0.943 & 0.951 & \textbf{0.964} & \textbf{0.980} & 0.944 \\
        ~ & LSFA \cite{tu2024self} & 0.957& 0.976& 0.970& 0.912& 0.934& 0.851& 0.960& 0.957& 0.970& 0.961& 0.945\\
        ~ & BridgeNet (Ours) & \textbf{0.972} & \textbf{0.976} & \textbf{0.982} & \textbf{0.920} & 0.978 & 0.871 & \textbf{0.978} & \textbf{0.960} & 0.961 & 0.956 & \textbf{0.955} \\
        \midrule
        \multirow{5}*{\rotatebox[origin=c]{90}{\textbf{3D+RGB}}} & EasyNet \cite{chen32} & 0.839 & 0.864 & 0.951 & 0.618 & 0.828 & 0.839 & 0.942 & 0.889 & 0.911 & 0.528 & 0.821 \\
        ~ & M3DM \cite{wang20} & 0.970 & 0.971 & 0.979 & 0.950 & 0.941 & 0.932 & 0.977 & 0.971 & 0.971 & 0.975 & 0.964 \\
        ~ & Shape-Guided \cite{chu21} & 0.981 & 0.973 & 0.982 & \textbf{0.971} & 0.962 & \textbf{0.978} & 0.981 & \textbf{0.983} & 0.974 & 0.975 & 0.976 \\
        ~ & 3DSR \cite{zavrtanik30} & 0.964 & 0.966 & 0.981 & 0.942 & \textbf{0.980} & 0.973 & 0.981 & 0.977 & \textbf{0.979} & 0.979 & 0.972 \\
        ~ & CFM \cite{costanzino35} & 0.980& 0.966& 0.982& 0.947& 0.959& 0.967& 0.982& \textbf{0.983}& 0.976& \textbf{0.982}& 0.972\\
        ~ & LSFA \cite{tu2024self} & \textbf{0.986}& 0.974& 0.981& 0.946& 0.925& 0.941& 0.983& 0.983& 0.974& 0.983& 0.968\\
        ~ & BridgeNet (Ours) & 0.978 & \textbf{0.979} & \textbf{0.984} & \textbf{0.971} & 0.978 & 0.973 & \textbf{0.983} & \textbf{0.983} & 0.976 & 0.967 & \textbf{0.977} \\
        \bottomrule
    \end{tabular}}
    \label{tab2}
\end{table*}

\subsection{Dual-modal Discriminator}
The dual-modal discrimination layer $F_{\varphi}$ is defined as:
\begin{equation}
    F_{\varphi}(\cdot)=\operatorname{Sigmod}(\operatorname{LeakyReLU}(\operatorname{BN}(\operatorname{Linear}(\cdot)))).
\end{equation}

\textbf{Training phase} The Dual-modal Discriminator is fed the final features from the three modules, outputs the predicted mask, and calculates loss using the corresponding ground truth mask. BCE LOSS is applied to the predicted masks of anomaly-free and Gaussian anomalous features, respectively, to help the model recognize the differences between anomalous and normal features. Focal LOSS \cite{ross31} is applied to the output of texture anomaly prediction masks to improve the robustness of the segmentation of anomalous products. The corresponding loss function is formulated as below:

The first term ${\mathcal{L}_{BCE}}_{n}$ is defined as normal dual-modal mask loss:
\begin{equation}
    {\mathcal{L}_{BCE}}_{n}\!=\!f_{B C E}(u,\!0)\!+\!f_{B C E}(u^{\prime},\!0)\!+\!f_{B C E}(u^{\prime\prime},\!0),
\end{equation}
where $u, u^{\prime}, u^{\prime \prime}$ are the predicted masks corresponding to the Gaussian anomaly concatenate at different scales. The second term ${\mathcal{L}_{BCE}}_{g}$ is defined as multi-scale Gaussian mask loss: 
\begin{equation}
    {\mathcal{L}_{BCE}}_{g}\!=\!f_{B C E}\left(u_{-},\!1\!\right)\!+\!f_{B C E}\left(u_{-}^{\prime},\!1\!\right)\!+\!f_{B C E}\left(u_{-}{\!}^{\prime\prime},\!1\!\right),
\end{equation}
where $u_{-}, u_{-}^{\prime}, u_{-}^{\prime \prime}$ are the predicted masks of Gaussian anomalies at different scales. The third term ${L_{FOCAL}}_{t}$ is defined as texture mask loss:
\begin{equation}
    {\mathcal{L}_{FOCAL}}_{t}=f_{F O C A L}\left(u_{+}, m_{t}\right),
\end{equation}
where $u_{+}$ is the predicted masks of texture anomalies. The total loss function for the training phase is represented as:
\begin{equation}
    \mathcal{L}={\mathcal{L}_{BCE}}_{n}+{\mathcal{L}_{BCE}}_{g}+{\mathcal{L}_{FOCAL}}_{t}.
\end{equation}

\textbf{Inference phase} The dual-modal discriminator is fed only test images and outputs anomaly scores $score_s$. Finally, the pixel-level anomaly score $score_l$ is obtained by linearly interpolation to upsample and Gaussian smoothing. The image-level anomaly score is the maximum of the pixel-level scores.

\begin{table*}[!t]
\centering
\caption{Comparison of BridgeNet with SOTA works on the MVTec-3D AD dataset in 5-shot, 10-shot, 50-shot, and full training set scenarios.}
\scalebox{1.0}{\begin{tabular}{l|cccc|cccc|cccc}
\hline
 & \textbf{5-shot} & \textbf{10-shot} & \textbf{50-shot} & \textbf{Full} & \textbf{5-shot} & \textbf{10-shot} & \textbf{50-shot} & \textbf{Full} & \textbf{5-shot} & \textbf{10-shot} & \textbf{50-shot} & \textbf{Full} \\
\hline
\textbf{Method} & \multicolumn{4}{c}{\textbf{I-AUROC}} & \multicolumn{4}{c}{\textbf{P-AUROC}} & \multicolumn{4}{c}{\textbf{P-AUPRO}} \\
\cline{1-13}
BTF \cite{horwitz18}  & 0.671 & 0.695 & 0.806 & 0.865 & 0.980 & 0.983 & 0.989 & 0.992 & 0.920 & 0.928 & 0.947 & 0.959 \\
AST \cite{rudolph19} & 0.680 & 0.689 & 0.794 & 0.937 & 0.950 & 0.946 & 0.974 & 0.976 & 0.903 & 0.835 & 0.929 & 0.944 \\
M3DM \cite{wang20} & 0.822 & 0.845 & 0.907 & 0.945 & 0.984 & 0.986 & 0.989 & 0.992 & 0.937 & 0.943 & 0.955 & 0.964 \\
CFM \cite{costanzino35}& 0.811 & 0.845 & 0.906 & 0.954 & 0.986 & 0.987 & 0.991 & 0.993 & \textbf{0.949} & 0.954 & 0.965 & 0.971 \\
LSFA \cite{tu2024self}& 0.834 & 0.871 & 0.926 & 0.971 & 0.984 & 0.987 & 0.989 & 0.993 & 0.936 & 0.943 & 0.962 & 0.968 \\
BridgeNet (Ours) & \textbf{0.883} & \textbf{0.911} & \textbf{0.977} & \textbf{0.993} & \textbf{0.988} & \textbf{0.993} & \textbf{0.996} & \textbf{0.996} & 0.946 & \textbf{0.955} & \textbf{0.972} & \textbf{0.977} \\
\hline
\end{tabular}}
\label{tabS4}
\end{table*}

\section{Experiment}
\textbf{Dataset} MVTec-3D AD \cite{bergmann17} contains 10 classes of 2D RGB images and aligned point cloud from manufacturing, 2,656 training samples, 294 validation samples, and 1,197 test samples. 
The training and validation sets have only normal samples, while the test data contains both normal and anomaly samples.
In addition to anomalies such as color, which are obvious in the RGB data, some anomalies, such as holes, can only be recognized from the depth. 

To validate the generalization of our method, we utilize an additional 3D anomaly detection dataset, Eyecandies \cite{bonfiglioli36}, which contains both RGB images and ordered point clouds. Each sample includes six RGB images captured under different lighting conditions and corresponding 3D ordered point clouds.

\textbf{Evaluation Metrics} As with most methods \cite{rudolph19, wang20, chu21, chen32}, we use three metrics to evaluate model performance. 
Image-level anomaly detection performance and pixel-level localization performance are evaluated by Area Under the Receiver Operator Curve (I-AUROC and P-AUROC). 
We use the per region overlap (P-AUPRO) for anomaly localization performance at pixel level.

\textbf{Data Preprocess} Since the point cloud in the MVTec-3D AD dataset is an aligned and ordered point cloud, for the 3D data, we fill in the missing values of the original depth image and separate the foreground and background based on the distance of the points to the background plane via \cite{zavrtanik30}. The foreground will be used to generate local anomaly masks for RGB and 3D data. We only use the depth component, \textit{i.e.}, z-dimensional values, and discard the x,y-dimensional values to generate depth images. 

For Eyecandies, we randomly select 400 training samples as the training data per epoch and convert them to the same data format as MVTec-3D AD, following the approach of M3DM \cite{wang20}. We use only the environment light RGB images as the input images. The remaining settings are consistent with the MVTec-3D AD dataset.

\textbf{Anomaly Generation} For dual-modal local anomaly generation, the degree of anomaly $\beta\sim\mathcal{N}\left(0.5, 0.3\right)$, truncated at (0.2, 0.8). The Gaussian noise generated by the dual-modal Gaussian anomaly follows $\mathcal{N}\left(0, 0.12\right)$, $\mathcal{N}\left(0, 0.06\right)$ and $\mathcal{N}\left(0, 0.02\right)$ Gaussian distribution, respectively. The parameter $\alpha$ in both texture and Gaussian anomaly combinations is set to 1/3.

\textbf{Model Parameter Setting} 
To reduce tuning burden and ensure generalizability, BridgeNet parameters like $\alpha$ and $\beta$ follow widely adopted empirical values from prior studies \cite{liu1}. 
We take WideResNet50, which has been pre-trained on the ImageNet dataset, as the backbone network of the Feature Extractor. We select 2,3 layers of features for subsequent inputs and set the feature dimension to 1536.
The Fusion Adaptor is a bias-free fully connected layer, with identical input and output dimensions and a learning rate of 0.00005.
Finally, for Discriminator, we set the input feature dimension to 1536, where the LeakyReLU activation uses a slope of 0.3 for negative values, the dimension of hidden layers is also set to 1536, and the learning rate is set to 0.0001. The training epochs are set to 160 and
the batch size is 4. The model was implemented with PyTorch 2.3.1, Python 3.9.13, and one NVIDIA Tesla V100.

\begin{figure}[h]
    \centering
    \includegraphics[width=0.88\linewidth]{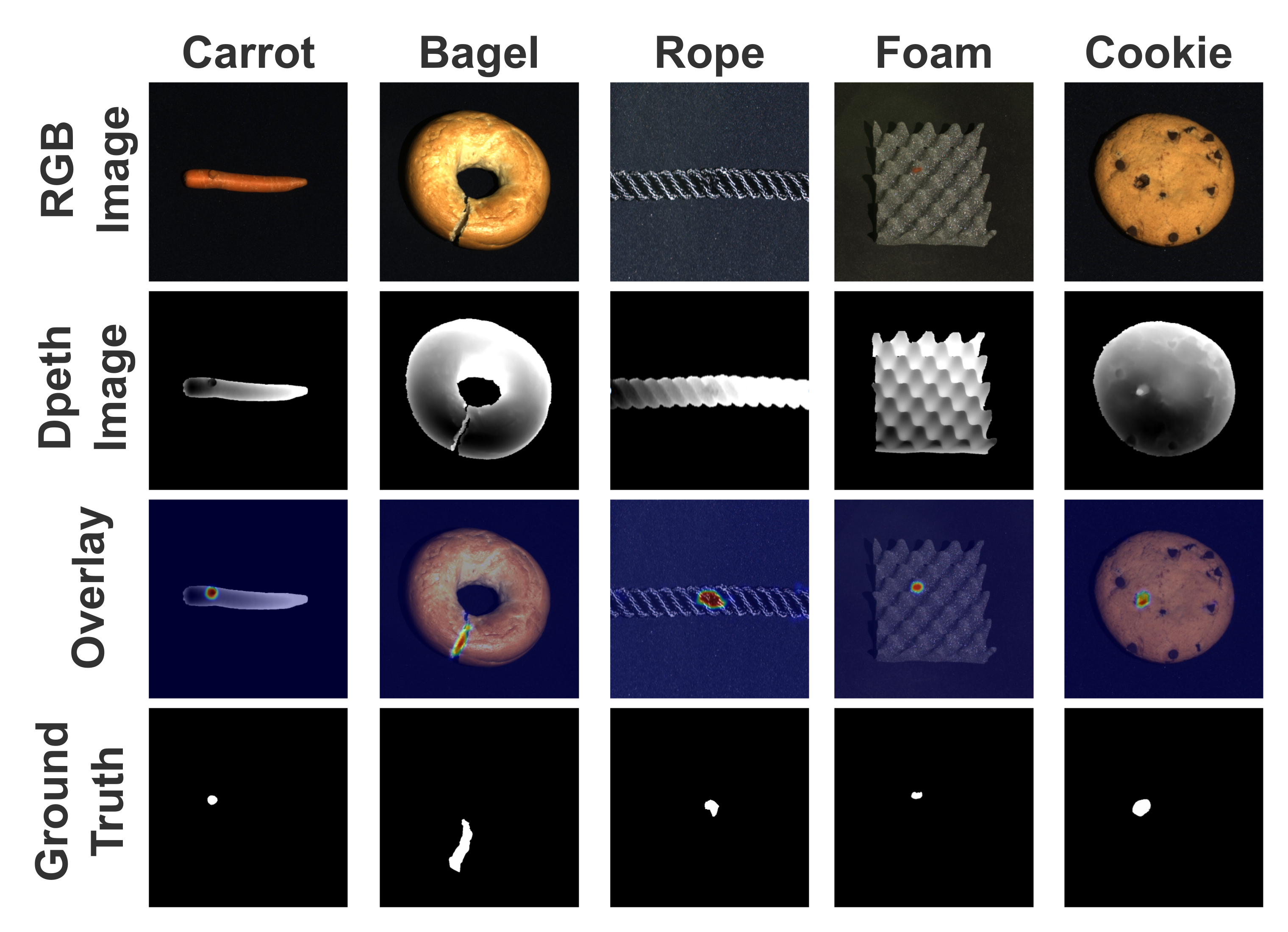}
    \caption{Qualitative results of BridgeNet on the MVTec-3D AD dataset.}
    \label{fig4}
\end{figure}

\subsection{Main Results}
\textbf{Main Results on MVTec-3D AD} BridgeNet is compared with SOTA works in 3D, RGB and 3D+RGB settings. For I-AUROC, Table \ref{tab1} shows that BridgeNet demonstrates new SOTA anomaly detection capability across three different settings, indicating the effectiveness of MGAG. In the 3D setting, BridgeNet with an average score 1.3\% higher than the second-best, 3DSR \cite{zavrtanik30}, indicating the feasibility of UTAG. In the 3D-RGB setting, BridgeNet ranks first in seven samples, with all samples being in the top two and an average score of 0.993, 1.5\% higher than the SOTA method 3DSR \cite{zavrtanik30}. This demonstrates that our share-weight framework can effectively learn the joint representation of RGB and depth information.

For P-AUPRO, Table \ref{tab2} demonstrates the SOTA anomaly localization capability of BridgeNet. BridgeNet achieved scores of 0.945, 0.955, and 0.977 in the three settings, all outperforming the SOTA methods ShapeGuided \cite{chu21} and 3DSR \cite{zavrtanik30}. 

Figure ~\ref{fig4} shows BridgeNet’s qualitative results on the MVTec-3D AD dataset: Rows 1–2 present RGB and depth images, Row 3 overlays the predicted mask on the RGB image, and Row 4 shows the ground truth. The Rope sample demonstrates the ability to detect subtle anomalies in both RGB and depth images. In contrast, the Foam sample highlights the effectiveness in locating anomalies that are clear in RGB but subtle in depth. Finally, the Cookie sample illustrates cases where anomalies are prominent in depth but subtle in RGB. All localization and segmentation results show that BridgeNet effectively identifies cross-modality anomaly combinations and delivers strong detection and segmentation performance.

\textbf{Few-shot Anomaly Detection} We also evaluated BridgeNet under few-shot setting and compared with recent SOTA few-shot models, as shown in Table \ref{tabS4}. We randomly selected 5, 10, and 50 samples from each class to serve as training data. Meanwhile, other settings were kept consistent with those in the main experiment. The experimental results show that BridgeNet achieves SOTA methods in detection and localization performance in few-shot settings. 

The appendix contains the few-shot evaluation results for all sample categories.

\textbf{Main Results on Eyecandies} We also use I-AUROC and P-AUPRO to evaluate detection and localization performance of our model. As in Table \ref{tabE1}, our method achieves SOTA via I-AUROC and competitive P-AUPRO results, showing SOTA detection and excellent localization capability on Eyecandies.

The appendix contains the evaluation results for all sample categories of Eyecandies.

\begin{table}[!t]
    \centering
    \caption{BridgeNet achieves superior I-AUROC scores for anomaly detection across all Eyecandies categories, outperforming SOTA works in 3D, RGB, and combined settings.}
    \scalebox{1.0}{\begin{tabular}{@{}llcc}
        \toprule
          & \multirow{2}*{\textbf{Method}} & \multirow{2}*{\textbf{I-AUROC}} &\multirow{2}*{\textbf{P-AUPRO}}\\
          & ~ & ~  &~  \\
        \midrule
        \multirow{7}*{\rotatebox[origin=c]{90}{\textbf{3D+RGB}}} & BTF \cite{horwitz18}& 0.821&0.846 \\
        ~ & EasyNet \cite{chen32} & 0.869 
 & - \\
        ~ & M3DM \cite{wang20} & 0.897
 &0.882
\\
        ~ & 3DSR \cite{zavrtanik30} & 0.909
 & - \\
        ~ & CFM\cite{costanzino35} & 0.881
 &0.887
\\
        ~ & LDM \cite{liu35}& 0.948
 &
\textbf{0.941}
\\
        ~ & BridgeNet (Ours) & \textbf{0.958} &0.929\\
        \bottomrule
    \end{tabular}}
    \label{tabE1}
\end{table}
\subsection{Ablation Study}

\textbf{Anomaly Generation Methods} Table \ref{tab7} investigates four dual-modal anomaly generators: (1) only UTAG, (2) only MGAG, (3) both without selective anomalies, (4) both with selective anomalies. Comparing rows 1, 2, 4, both generators are crucial for training the dual-modal discriminator, with MGAG more vital for enhancing anomaly segmentation. Comparing rows 3–4, selective modalities on both generators improve detection and localization performance.

The appendix contains additional experimental results on the discussion of the generalizability of MGAG and UTAG.
\begin{table}[!t]
    \centering
    \caption{Performance comparison of different anomaly generators combination.}
    \scalebox{1.0}{\begin{tabular}{@{}lccc@{}}
        \toprule
        \textbf{Method} & \textbf{I-AUROC} & \textbf{P-AUROC} & \textbf{P-AUPRO} \\
        \midrule
        UTAG & 0.946& 0.983& 0.949\\
        MGAG & 0.969& \textbf{0.996}& 0.965\\
        UTAG+MGAG ($\alpha=0$) & 0.992& \textbf{0.996}& 0.975\\
        UTAG+MGAG ($\alpha=1/3$) & \textbf{0.993}& \textbf{0.996}& \textbf{0.977}\\
        \bottomrule
    \end{tabular}}
    \label{tab7}
\end{table}
\begin{table}[!t]
    \centering
    \caption{Impact of Gaussian noise injection at different layers on anomaly detection and localization performance.}
    \scalebox{1.0}{\begin{tabular}{@{}cccccc@{}}
        \toprule
        \textbf{G1} & \textbf{G2} & \textbf{G3} & \textbf{I-AUROC} & \textbf{P-AUROC} & \textbf{P-AUPRO} \\
        \midrule
        \checkmark & & & 0.893& 0.953& 0.867\\
        & \checkmark & & 0.988& 0.995& 0.968\\
        & & \checkmark & 0.988& 0.995& 0.970\\
        \checkmark & \checkmark & & 0.989& 0.995& 0.972\\
        \checkmark & & \checkmark & 0.987& 0.995& 0.969\\
        & \checkmark & \checkmark & 0.992& \textbf{0.996}& 0.974\\
        \checkmark & \checkmark & \checkmark & \textbf{0.993}& \textbf{0.996}& \textbf{0.977}\\
        \bottomrule
    \end{tabular}}
    \label{tab:multi_scale_gaussian_noise}
\end{table}

\textbf{Gaussian Anomaly Generation in Different Layers} Table \ref{tab:multi_scale_gaussian_noise} compares how adding Gaussian anomalies to different model parts affects results. From the first three rows, adding noise solely at G2 and G3 yields excellent results, while doing so at G1 is less effective. The reason may be that when noise perturbs at the image level, it propagates to deeper features, creating an excessive variety of feature anomalies, making it difficult for the model to distinguish real anomalies effectively. When combining G1 with G2 and G3 anomalies, as well as all, the model achieves results surpassing the current SOTA performance. The possible reason is that in the case of combined anomalies, the Gaussian anomalies at the feature level already assist the model in identifying anomalies adequately, and the added Gaussian noise at G1 further enriches the anomalies at the feature level, which enhances the robustness of the discrimination layer. The table shows (G1, G2, G3) improves I-AUROC and P-AUPRO by 0.1\% and 0.3\% compared to (G2, G3). Similarly, (G1, G2) increases them by 0.1\% and 0.4\% versus G2 alone.

\begin{figure} [!t]
    \centering
    \subfloat[Rope Feature Std]{\includegraphics[scale=0.12]{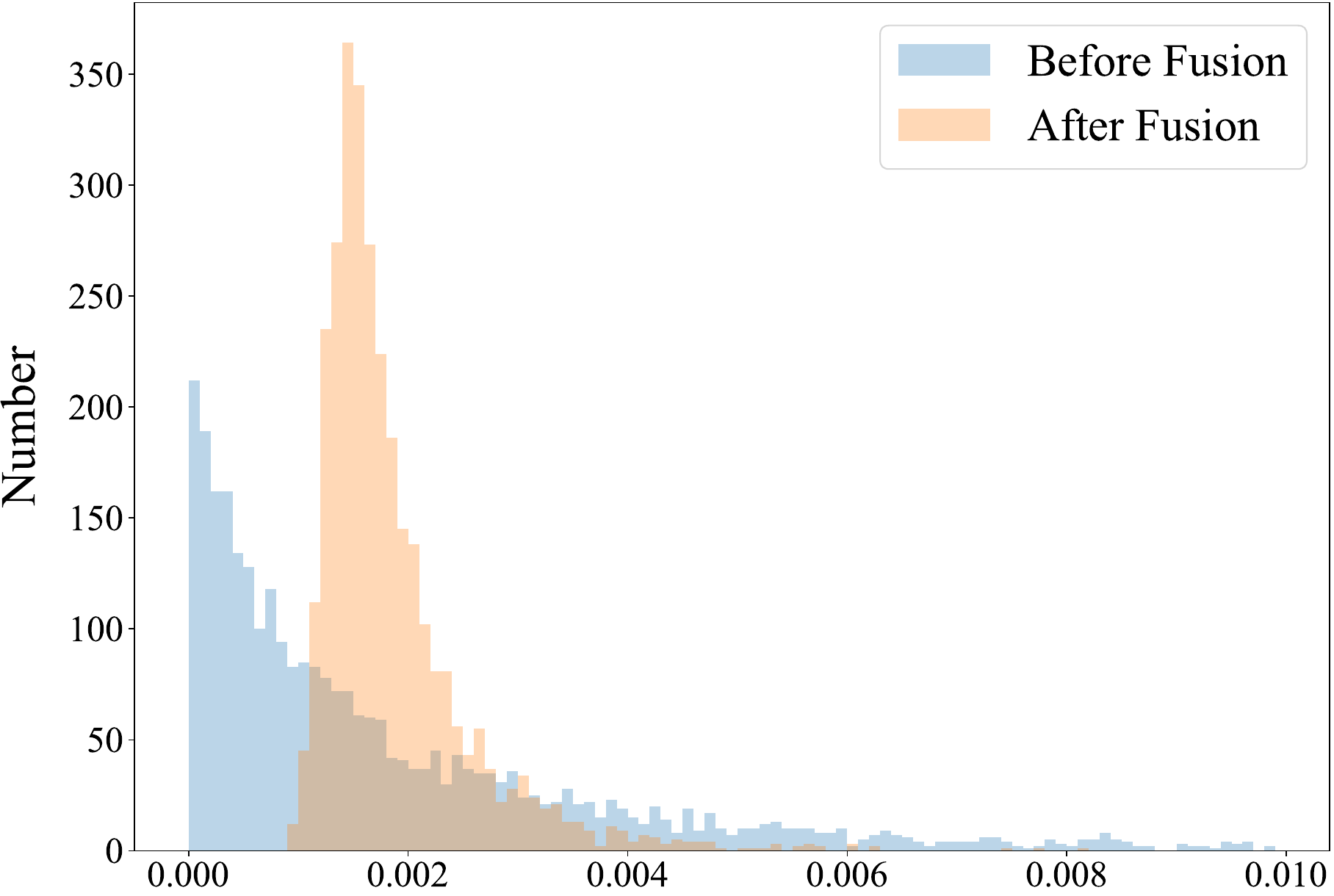}}
    \subfloat[Foam Feature Std]{\includegraphics[scale=0.12]{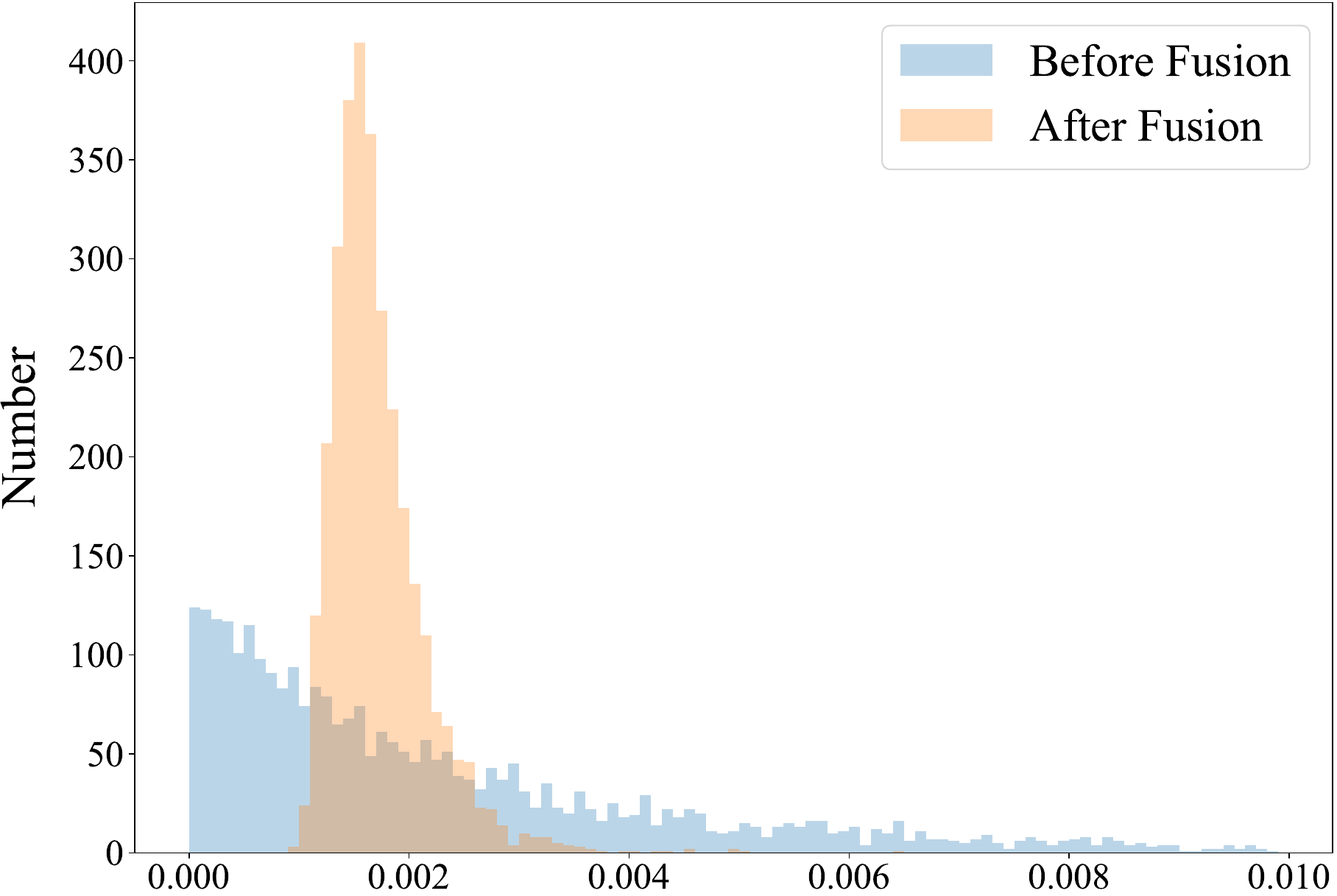}}
    \caption{Histograms of standard deviations per dimension for local and adapted features of (a) Rope and (b) Foam.}
    \label{fig7} 
\end{figure}

\textbf{Gaussian Anomaly Generation at Different Scales} Table \ref{tab:multi_scale_noise} further explores how the scale of Gaussian noise added at different layers impacts anomaly detection and localization results. Here, $\sigma_{1}$, $\sigma_{2}$, and $\sigma_{3}$ correspond to the scales of Gaussian noise at G1, G2, and G3, respectively. Adding noise of the same scale at all three layers did not yield ideal results. Better results are often achieved when noise scales satisfy $\sigma_1 > \sigma_2 > \sigma_3$, as shown in rows 5 and 6. The reason is that the feature space becomes more compact after passing through the Fusion Adaptor, as shown in Figure ~\ref{fig7}. Therefore, larger-scale noise needs to be added to features before transformation than after, to ensure anomaly signals propagate sufficiently and are captured by the model. As shown in rows 4 and 5, adding larger-scale Gaussian noise at G2 improves all three overall metrics by 0.5\%, 0.1\%, and 0.6\%, respectively. Additionally, since the value range at the image level often spans more widely, relatively larger perturbations are needed to generate sufficient anomaly signals. As shown in rows 6 and 7, increasing the noise scale enhanced I-AUROC and P-AUPRO by 2.1\% and 1.1\%, respectively.

\textbf{Different methods in multimodal scenarios with parameter-sharing framework} Both AST+ and CFM+ follow BridgeNet's approach for 3D information preprocessing. They convert raw point clouds into depth images, extract features using the same 2D pre-trained model, and concatenate them to replace subsequent model inputs. GLASS-3D replaces BridgeNet's Multi-scale Gaussian Anomaly Generator with a Controllable Gaussian Anomaly Generator while keeping the same texture anomaly generation method. Results show that using identical pre-trained models in multimodal scenarios enhances the performance of embedding-based anomaly detection methods. Notably, CFM+ achieves this even with reduced feature dimensions (1152$\to$768) after replacing PointMAE with DINO\_ViTBase8. Meanwhile, although GLASS excels in 2D scenarios (SOTA performance), its effectiveness diminishes slightly in multimodal settings. This is due to the increased diversity of anomalies during inference compared to our method.

\begin{table}[!t]
    \centering
    \caption{Impact of different Gaussian noise scales on anomaly detection and localization performance.}
    \scalebox{1.0}{
        \begin{tabular}{@{}cccccc@{}}
            \toprule
            $\sigma_{1}$ & $\sigma_{2}$ & $\sigma_{3}$ & \textbf{I-AUROC} & \textbf{P-AUROC} & \textbf{P-AUPRO} \\
            \midrule
            0.02 & 0.02 & 0.02 & 0.977& 0.994& 0.962\\
            0.04 & 0.04 & 0.04 & 0.981& 0.994& 0.965\\
            0.12 & 0.12 & 0.12 & 0.941& 0.991& 0.953\\
            $\times$ & 0.02 & 0.02 & 0.987 & 0.995& 0.968
\\
            $\times$ & 0.04 & 0.02 & 0.992 & \textbf{0.996}& 0.974
\\
            0.02 & $\times$ & 0.02 & 0.966& 0.995& 0.958\\
            0.12 & $\times$ & 0.02 & 0.987 & 0.995& 0.969
\\
            0.04 & 0.03 & 0.02 & 0.991 & 0.995& 0.972
\\
            0.12 & 0.04 & 0.02 & \textbf{0.993}& \textbf{0.996}& \textbf{0.977}\\
            \bottomrule
        \end{tabular}
    }
    \label{tab:multi_scale_noise}
\end{table}
\begin{table}[!t]
    \centering
    \caption{Performance comparison of different methods with sharing weight model in multimodal anomaly detection.}
    \scalebox{1.0}{\begin{tabular}{@{}lccc@{}}
        \toprule
        \textbf{Method} & \textbf{I-AUROC} & \textbf{P-AUROC} & \textbf{P-AUPRO} \\
        \midrule
        AST \cite{rudolph19} & 0.937& 0.976& 0.944\\
        AST+ & 0.972& 0.990& 0.956\\
        CFM \cite{costanzino35} & 0.960& 0.993& 0.972\\
        CFM+ & 0.966& 0.994& 0.973\\   
        GLASS-3D & 0.989& 0.995& 0.973\\
        BridgeNet (Ours) & 0.993& 0.996& 0.977\\
        \bottomrule
    \end{tabular}}
    \label{tab8}
\end{table}
\section{Conclusion}
We propose a novel 3D anomaly detection method, BridgeNet, which effectively bridges 2D and 3D information. Specifically, BridgeNet shares all parameters in model between the two modalities, achieving implicit alignment of dual-modal features. Leveraging similar feature distributions for both modalities, we provide a unified texture anomaly generator and multi-scale Gaussian anomaly generator, generating richer Gaussian of different semantic levels and unified texture anomalies. BridgeNet achieves new SOTA results on the MVTec-3D AD and Eyecandies datasets, and we hope it can inspire future research in this field.

\section{Acknowledgments}
This work is supported by the National Natural Science Foundation of China (92267105), 
Guangdong Basic and Applied Basic Research Foundation (2023B1515130002), 
Guangdong Special Support Plan (2021TQ06X990), 
Shenzhen Basic Research Program (JCYJ20220818101610023, KJZD20230923113800001), 
National Natural Science Foundation of China (62376263),
Natural Science Foundation of Guangdong (2024A1515030209),
Shenzhen Science and Technology Innovation Commission (JCYJ20230807140507015).

\bibliographystyle{ACM-Reference-Format}
\balance  
\bibliography{sample-base}

\newpage
\appendix

\begin{table}[!t]
    \centering
    \caption{Impact of different Gaussian anomaly generation method on Visa dataset.}
    \scalebox{1}{\begin{tabular}{@{}lccc@{}}
        \toprule
        \textbf{Method} & \textbf{I-AUROC} & \textbf{P-AUROC} \\
        \midrule
        SimpleNet \cite{liu1} & 0.980& 0.985\\
        MGAG & 0.989& 0.991\\
        GLASS \cite{chen16} & 0.988& 0.988\\
        GLASS \cite{chen16} + MGAG & \textbf{0.992} & \textbf{0.992}\\

        \bottomrule
    \end{tabular}}
    \label{tab_2d_visa}
\end{table}

\begin{table}[!t]
    \centering
    \caption{UTAG vs. Perlin Noise-based Anomaly Generation Strategy.}
    \scalebox{1}{\begin{tabular}{@{}lccc@{}}
        \toprule
        \textbf{Method} & \textbf{I-AUROC} & \textbf{P-AUPRO} \\
        \midrule
        EasyNet \cite{chen32} & 0.926& 0.821\\
        EasyNet (With UTAG) & 0.941& 0.859\\
        BridgeNet (With Perlin Noise) & 0.984& 0.970\\
        BridgeNet (Ours) & \textbf{0.993} & \textbf{0.977}\\

        \bottomrule
    \end{tabular}}
    \label{tab_utag}
\end{table}

\begin{table}
\centering
\caption{The table shows that BridgeNet balances high
accuracy with efficient memory usage and FPS.}
\begin{tabular}{lccccccccccc}
\hline
\textbf{Method} & \textbf{CFM} & \textbf{Shape-Guided} & \textbf{M3DM} & \textbf{3DSR} & \textbf{Ours} \\
\hline
Memory(MB) & \textbf{621} & 6155& 11962& 2921& 2289
\\
FPS & 21 & 0.3& 0.7& \textbf{38} & 25
\\

\hline
\end{tabular}
\label{tab_fps}
\end{table}

\begin{table}[!t]
    \centering
    \caption{Impact of different Gaussian anomaly generation method on MVTec AD dataset.}
    \scalebox{1}{\begin{tabular}{@{}lccc@{}}
        \toprule
        \textbf{Method} & \textbf{I-AUROC} & \textbf{P-AUROC} \\
        \midrule
        SimpleNet \cite{liu1} & 0.998& 0.988\\
        MGAG & \textbf{0.999}& 0.990\\
        GLASS \cite{chen16} & \textbf{0.999}& 0.993\\
        GLASS \cite{chen16} + MGAG & \textbf{0.999}& \textbf{0.994}\\

        \bottomrule
    \end{tabular}}
    \label{tab_2d_mvtec}
\end{table}

\section{Overview}

This document is structured as follows: 
\begin{itemize}
    \item \ref{DR}: This section presents the I-AUROC and P-AUPRO metrics for each category in one-shot and few-shot scenarios, along with a comparison to the SOTA few-shot methods.

    \item \ref{ER}: This section describes the experimental setup and detailed results analysis of our method on the Eyecandies dataset.

    \item \ref{C2}: This section additionally explores the performance of our MGAG module solely on 2D datasets.

    \item \ref{UT}: This section compares UTAG with anomaly generation methods based on Perlin noise.

    \item \ref{ME}: This section presents the memory usage and FPS of our model during the inference stage.

    \item \ref{VR}: This section shows the results of additional qualitative analysis on the MVTec-3D AD and Eyecandies datasets.

\end{itemize}

\section{Detailed Results of One-shot and Few-shot scenarios}
\label{DR}

\textbf{Experimental setup} In the one-shot and few-shot scenarios, we randomly select the corresponding number of samples from each category of the MVTec-3D AD dataset and replicate them 200 times to form the training data for each epoch. The full test dataset is then used to evaluate our model.

\textbf{Results Analysis}: As shown in Table \ref{tabS3} and Table \ref{tabS5}, we present the I-AUROC and P-AUROC metrics in the MVTec-3D AD dataset in the 1-shot, 5-shot, 10-shot, and 50-shot scenarios. 
Our method performs poorly in the 1-shot scenario due to the test samples in certain categories, such as carrot, cable, and peach, having diverse orientations and significant variations.
This makes it challenging for the model to learn sufficient information from just a single sample. Therefore, applying data augmentation through rotations in the one-shot scenario may improve results. 
In the 5-shot and 10-shot scenarios, both the I-AUROC and P-AUROC metrics show significant improvement, with the average results even outperforming some 3D anomaly detection methods by using the full dataset.
In the 50-shot scenario, the model achieves excellent results, matching the P-AUPRO performance of SOTA methods and the I-AUROC metric only 0.1\% lower.

\begin{figure}[!t]
    \centering
    \includegraphics[width=0.9\linewidth]{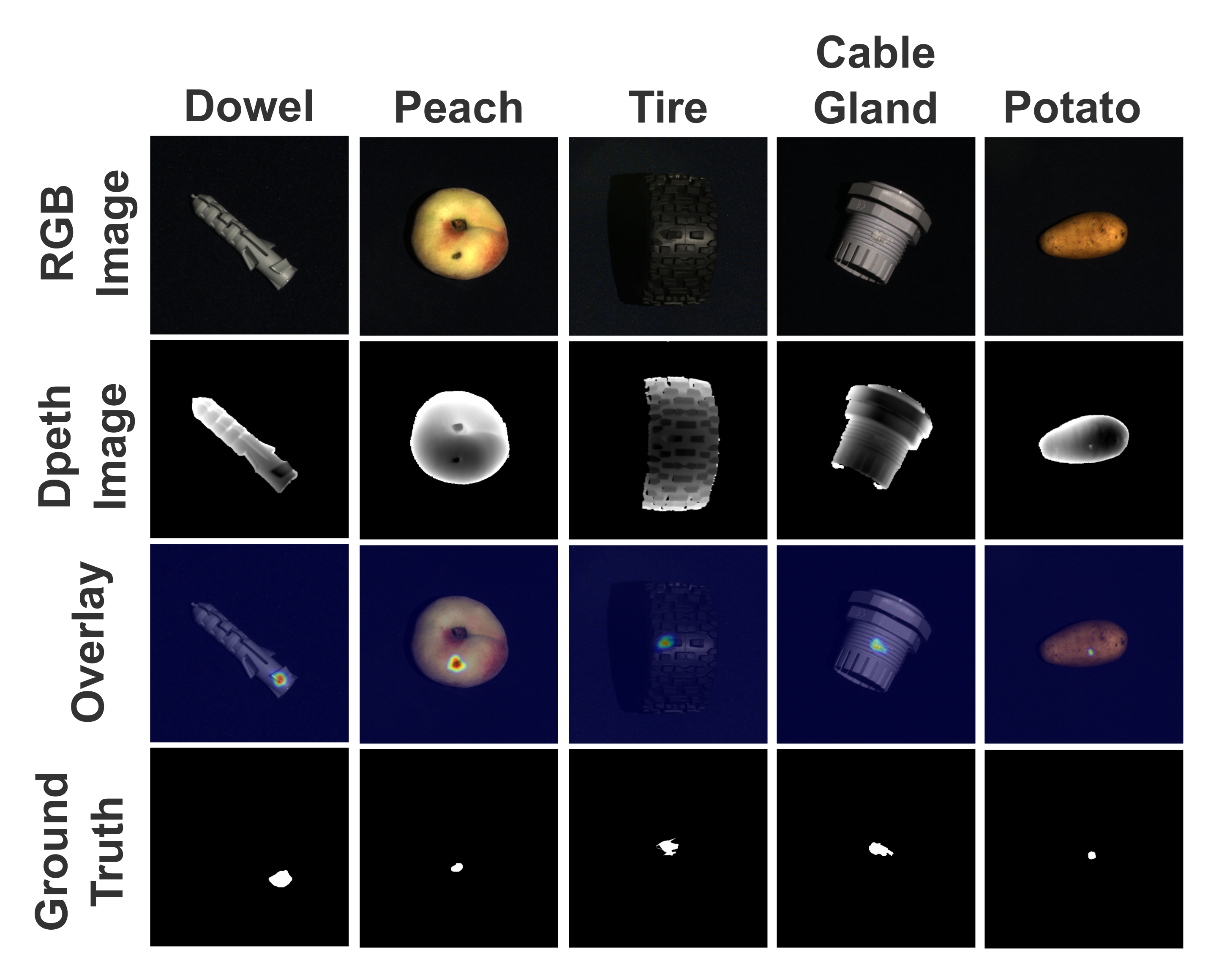}
    \caption{Qualitative results of BridgeNet on the MVTec-3D AD dataset.}
    \label{figs1}
\end{figure}

\begin{figure*}[!htbp]
    \centering
    \includegraphics[width=0.9\linewidth]{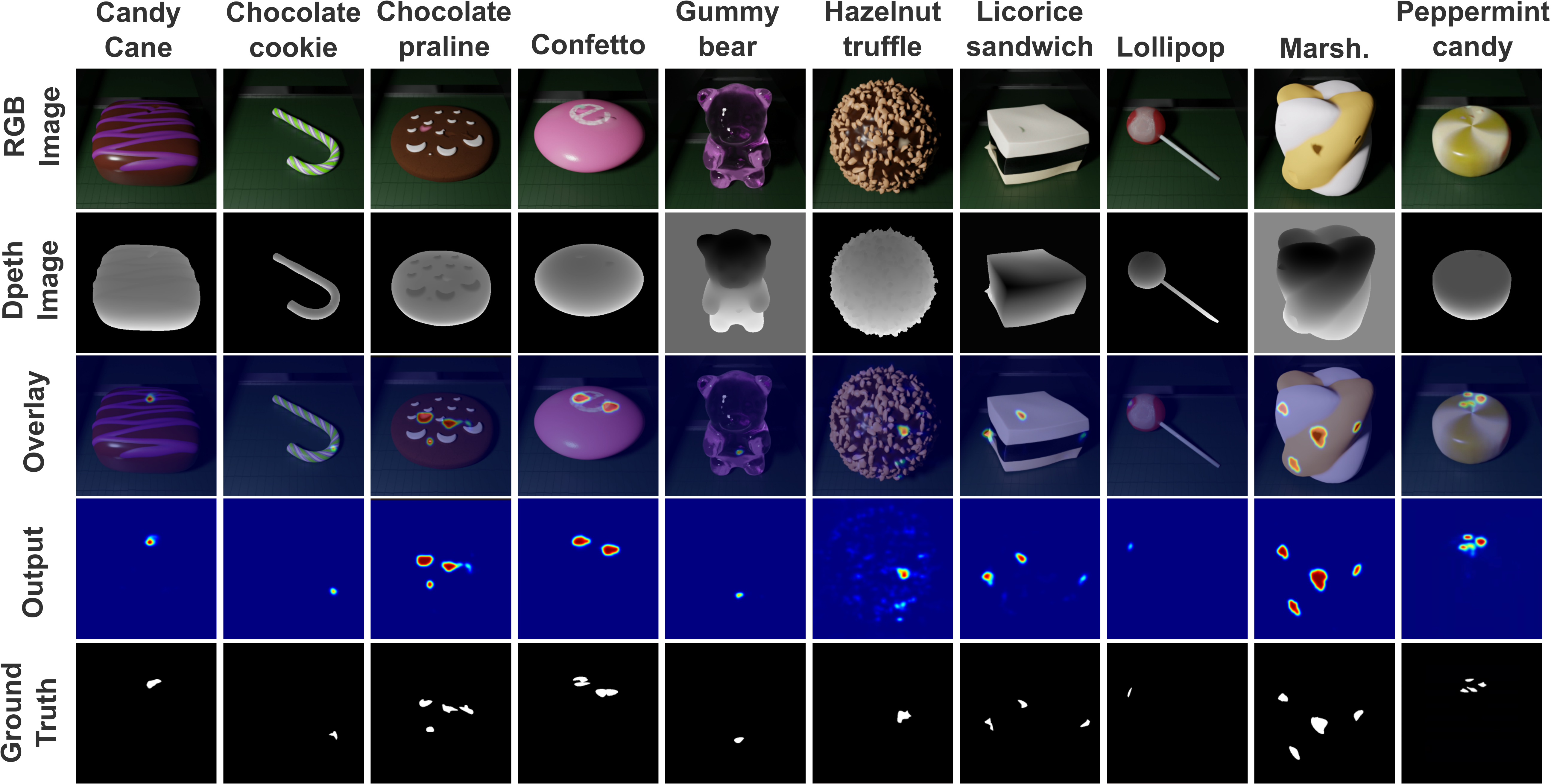}
    \caption{Qualitative results of BridgeNet on the Eyecandies dataset.}
    \label{figs2}
\end{figure*}

\begin{table*}
\centering
\caption{I-AUROC of BridgeNet on the MVTec-3D AD dataset in 1-shot, 5-shot, 10-shot and 50-shot scenarios.}
\begin{tabular}{lccccccccccc}
\hline
\textbf{Method} & \textbf{Bagel} & \textbf{Cable} \textbf{Gland} & \textbf{Carrot} & \textbf{Cookie} & \textbf{Dowel} & \textbf{Foam} & \textbf{Peach} & \textbf{Potato} & \textbf{Rope} & \textbf{Tire} & \textbf{Mean} \\
\hline
1-shot & 0.586& 0.523& 0.591& 0.783& 0.631& 0.807& 0.543& 0.788& 0.791& 0.670& 0.671
\\
5-shot & 0.986& 0.644& 0.945& 0.970& 0.793& 0.897& 0.905& 0.867& 0.964& 0.863& 0.883
\\
10-shot & 0.989& 0.767& 0.960& 0.962& 0.832& 0.912& 0.924& 0.880& 0.992& 0.891& 0.911
\\
50-shot & 1.000& 0.942& 0.970& 0.979& 0.989& 0.981& 0.988& 0.977& 0.999& 0.942& 0.977\\
\hline
\end{tabular}
\label{tabS3}
\end{table*}

\begin{table*}
\centering
\caption{P-AUPRO of BridgeNet on the MVTec-3D AD dataset in 1-shot, 5-shot, 10-shot and 50-shot scenarios.}
\begin{tabular}{lccccccccccc}
\hline
\textbf{Method} & \textbf{Bagel} & \textbf{Cable} \textbf{Gland} & \textbf{Carrot} & \textbf{Cookie} & \textbf{Dowel} & \textbf{Foam} & \textbf{Peach} & \textbf{Potato} & \textbf{Rope} & \textbf{Tire} & \textbf{Mean} \\
\hline
1-shot & 0.781& 0.615& 0.945& 0.883& 0.855& 0.880& 0.842& 0.938& 0.935& 0.799& 0.847
\\
5-shot & 0.955& 0.887& 0.971& 0.955& 0.923& 0.945& 0.964& 0.980& 0.952& 0.931& 0.946
\\
10-shot & 0.958& 0.932& 0.981& 0.949& 0.937& 0.952& 0.973& 0.983& 0.960& 0.926& 0.955
\\
50-shot & 0.972& 0.973& 0.982& 0.970& 0.973& 0.961& 0.982& 0.983& 0.962& 0.965& 0.972\\
\hline
\end{tabular}
\label{tabS5}
\end{table*}

\begin{table*}[!htbp]
    \centering
    \caption{BridgeNet achieves superior I-AUROC scores for anomaly detection across all Eyecandies categories, outperforming SOTA works in 3D, RGB, and combined settings.}
    \scalebox{0.89}{\begin{tabular}{@{}llccccccccccc@{}}
        \toprule
          & \multirow{2}*{\textbf{Method}} & \textbf{Candy} & \textbf{Chocolate}&\textbf{Chocolate}& \multirow{2}*{\textbf{Confetto}} & \textbf{Gummy}& \textbf{Hazelnut}& \textbf{Licorice}& \multirow{2}*{\textbf{Lollipop}}& \multirow{2}*{\textbf{Marsh.}}& \textbf{Peppermint}& \multirow{2}*{\textbf{Mean}} \\
          & ~ & \textbf{Cane} & \textbf{Cookie}&\textbf{Praline}& ~ & \textbf{Bear}& \textbf{Truffle}& \textbf{Sandwich}& ~ & ~ & \textbf{Candy}& ~ \\
        \midrule
        \multirow{6}*{\rotatebox[origin=c]{90}{\textbf{3D}}} & FPFH \cite{horwitz18}& 0.670& 0.728& 0.806& 0.806& 0.721& 0.514& 0.794& 0.757& 0.758& 0.757& 0.731
\\
        ~ & EasyNet \cite{chen32} & 0.629 & 0.716& 0.768& 0.731& 0.660& 0.710& 0.712& 0.711& 0.688& 0.731& 0.706 
\\
        ~ & M3DM \cite{wang20} & 0.482 & 0.589& 0.805& 0.845& 0.780& 0.538& 0.766& 0.827& 0.800& 0.822& 0.725
\\
        ~ & 3DSR \cite{zavrtanik30} & 0.600 & 0.768& 0.742 & 0.770& 0.761& \textbf{0.749}& 0.811& 0.831& 0.811& 0.917& 0.776
\\
        ~ & LDM \cite{liu35}& 0.790& \textbf{0.885}& \textbf{0.933}& \textbf{0.915}& \textbf{0.837}& 0.517& \textbf{0.888}& \textbf{0.960}& \textbf{0.941}& \textbf{0.949}& \textbf{0.862}
\\
        ~ & Ours & \textbf{0.912}& 0.846& 0.820& 0.827& 0.727& 0.611& 0.870& 0.921& 0.905& 0.878& 0.832\\
        \midrule
        \multirow{6}*{\rotatebox[origin=c]{90}{\textbf{RGB}}} & PatchCore \cite{horwitz18}& 0.522& 0.853& 0.621& 0.950& 0.710& 0.624& 0.866& 0.779& 0.982& 0.845& 0.775
\\
        ~ & EasyNet \cite{chen32} 
& 0.723  & 0.925& 0.849& 0.966& 0.705& \textbf{0.815}& 0.806& 0.851& 0.975& 0.960& 0.858
\\
        ~ & M3DM \cite{wang20} 
& 0.648 & 0.949& 0.941& \textbf{1.000}& \textbf{0.878}& 0.632& 0.933& 0.811& 0.998& \textbf{1.000}& 0.879
\\
        ~ & 3DSR \cite{zavrtanik30} 
& 0.706& 0.965& 0.950& 0.966& 0.870& 0.790& 0.885& 0.857& 0.998& 0.992& 0.898
\\
        ~ & LDM \cite{liu35}
& 0.710& \textbf{0.998}& \textbf{0.955}& 0.966& 0.843& 0.592& 0.947& 0.875& \textbf{1.000}& 0.995& 0.888
\\
        ~ & Ours & \textbf{0.837} & 0.979& 0.859& 0.989& 0.780& 0.727& \textbf{0.978}& \textbf{0.930}& \textbf{1.000}& \textbf{1.000}& \textbf{0.908}\\
        \midrule
        \multirow{7}*{\rotatebox[origin=c]{90}{\textbf{3D+RGB}}} & BTF \cite{horwitz18}& 0.712 & 0.882& 0.784& 0.942& 0.774& 0.579& 0.829& 0.840& 0.986& 0.882& 0.821
\\
        ~ & EasyNet \cite{chen32} & 0.737 & 0.934& 0.866& 0.966& 0.717& 0.822& 0.847& 0.863& 0.977& 0.960& 0.869
\\
        ~ & M3DM \cite{wang20} & 0.624 & 0.958& 0.958& \textbf{1.000}& 0.886& 0.758& 0.949& 0.836& \textbf{1.000}& \textbf{1.000}& 0.897
\\
        ~ & 3DSR \cite{zavrtanik30} & 0.651 & 0.998& 0.904& 0.978& 0.875& \textbf{0.861}& 0.965& 0.899& 0.990& 0.971& 0.909
\\
        ~ & CFM\cite{costanzino35} & 0.680 & 0.931& 0.952& 0.880& 0.865& 0.782& 0.917& 0.840& 0.998& 0.962& 0.881
\\
        ~ & LDM \cite{liu35}& 0.859& \textbf{1.000}& \textbf{1.000}& 0.995& 0.910& 0.738& \textbf{0.998}& \textbf{0.976}& \textbf{1.000}& \textbf{1.000}& 0.948
\\
        ~ & Ours & \textbf{0.960} & \textbf{1.000}& 0.965& 0.997& \textbf{0.912}& 0.822& 0.984& 0.941& \textbf{1.000}& \textbf{1.000}& \textbf{0.958}\\
        \bottomrule
    \end{tabular}}
    \label{tabS1}
\end{table*}

\begin{table*}[h]
    \centering
    \caption{BridgeNet demonstrates competitive P-AUPRO scores for anomaly localization across all Eyecandies categories, performing comparably to SOTA works in 3D, RGB, and combined settings.}
    \scalebox{0.89}{\begin{tabular}{@{}llccccccccccc@{}}
        \toprule
          & \multirow{2}*{\textbf{Method}} & \textbf{Candy} & \textbf{Chocolate}&\textbf{Chocolate}& \multirow{2}*{\textbf{Confetto}} & \textbf{Gummy}& \textbf{Hazelnut}& \textbf{Licorice}& \multirow{2}*{\textbf{Lollipop}}& \multirow{2}*{\textbf{Marsh.}}& \textbf{Peppermint}& \multirow{2}*{\textbf{Mean}} \\
          & ~ & \textbf{Cane} & \textbf{Cookie}&\textbf{Praline}& ~ & \textbf{Bear}& \textbf{Truffle}& \textbf{Sandwich}& ~ & ~ & \textbf{Candy}& ~ \\
        \midrule
        \multirow{4}*{\rotatebox[origin=c]{90}{\textbf{3D}}} & FPFH \cite{horwitz18}& \textbf{0.944}& 0.725& 0.687& 0.601& 0.651& 0.471& 0.636& 0.885& 0.598& 0.594& 0.679
\\
        ~ & M3DM \cite{wang20} & 0.911 & 0.645& 0.581& 0.748& 0.748& 0.484& 0.608& 0.904& 0.646& 0.750& 0.702
\\
        ~ & LDM \cite{liu35}& 0.842 & \textbf{0.841}& \textbf{0.870}& \textbf{0.868}& \textbf{0.814}& \textbf{0.591}& \textbf{0.838}& 0.865& \textbf{0.776}& 0.774& \textbf{0.808}
\\
        ~ & Ours & \textbf{0.944} & 0.768& 0.730& 0.820& 0.713& 0.543& 0.712& \textbf{0.925}& 0.667& \textbf{0.835}& 0.766\\
        \midrule
        \multirow{4}*{\rotatebox[origin=c]{90}{\textbf{RGB}}} & PatchCore \cite{horwitz18}& 0.773& 0.857& 0.594& 0.965& 0.762& 0.532& 0.887& 0.871& 0.942& 0.898& 0.808
\\
        ~ & M3DM \cite{wang20} 
& 0.867 & 0.904& 0.805& 0.982& 0.871& \textbf{0.662}& 0.882& 0.895& 0.970& 0.962& 0.880
\\
        ~ & LDM \cite{liu35}
& 0.919& 0.942& \textbf{0.887}& \textbf{0.978}& \textbf{0.910}& 0.627& \textbf{0.961}& \textbf{0.946}& \textbf{0.982}& \textbf{0.983}& \textbf{0.914}
\\
        ~ & Ours & \textbf{0.944}& \textbf{0.947}& 0.734& \textbf{0.978}& 0.854& 0.573& 0.885& 0.945& 0.968& 0.978& 0.881\\
        \midrule
        \multirow{5}*{\rotatebox[origin=c]{90}{\textbf{3D+RGB}}} & BTF \cite{horwitz18}& 0.871& 0.900& 0.698& 0.966& 0.823& 0.567& 0.884& 0.905& 0.953& 0.897& 0.846 
\\
        ~ & M3DM \cite{wang20} & 0.906& 0.826& 0.803& \textbf{0.983}& 0.855& 0.688& 0.880& 0.906& 0.966& 0.955& 0.882
\\
        ~ & CFM\cite{costanzino35} & 0.942& 0.902& 0.831& 0.965& 0.875& 0.762& 0.791& 0.913& 0.939& 0.949& 0.887
\\
        ~ & LDM \cite{liu35}& 0.964& 0.953& \textbf{0.951}& 0.982& \textbf{0.931}& \textbf{0.765}& \textbf{0.969}& 0.935& 0.982& \textbf{0.983}& \textbf{0.941}
\\
        ~ & Ours & \textbf{0.969} & \textbf{0.965}& 0.927& 0.978& 0.892& 0.681& 0.966& \textbf{0.946}& \textbf{0.983}& \textbf{0.983}& 0.929\\
        \bottomrule
    \end{tabular}}
    \label{tabS2}
\end{table*}

\section{Eyecandies Results} 
\label{ER}

\textbf{Results Analysis}: We also details the detection and localization performance for all
sample categories in Table ~\ref{tabS1} and Table ~\ref{tabS2}, our method achieves SOTA performance based on the I-AUROC metric in both the RGB+3D and RGB settings. Additionally, our model attains competitive results based on the P-AUPRO metric, demonstrating that our approach exhibits SOTA anomaly detection capability and excellent anomaly localization performance on the Eyecandies dataset.

\section{The performance of MGAG on 2D datasets} 
\label{C2}
Besides multi-modal scenarios, we also explored the role of MGAG in 2D datasets MVTec AD and Visa ~\cite{bergmann2019mvtec,zou2022spot}.

\textbf{Experimental settings}: In the experiments presented in Table ~\ref{tab_2d_mvtec} and Table \ref{tab_2d_visa} , all methods employed the same texture anomaly generation strategy as GLASS.
Specifically, for SimpleNet, Gaussian anomalies were added only after the Adapter layer. For MGAG, Gaussian anomalies of different scales were added at different positions in the model. In the case of GLASS+MGAG, the controllable Gaussian anomalies after the Adapter were maintained, and the Gaussian anomalies generated at other positions were consistent with those in MGAG. In addition to the above settings, the structural parameters of the overall model are the same as those of SimpleNet, and the number of training epochs is 640 for all.

\textbf{Results Analysis}: We found that the Gaussian anomalies of different scales could also help the model identify the disturbances caused by anomalies at different positions in 2D scenarios.

\section{Comparison of UTAG vs. Perlin Noise-based Anomaly Generation Methods} 
\label{UT}

\textbf{Experimental settings}: We first used the Perlin Noise-based anomaly generation method from EasyNet ~\cite{chen32} for 3D texture anomaly generation. Then we substituted it with our UTAG module in the same framework.

\textbf{Results Analysis}: Through comparison in Table ~\ref{tab_utag}, we find that UTAG demonstrates greater effectiveness than the Perlin Noise-based anomaly generation method in both our BridgeNet and EasyNet. Moreover, it indicates that 2D anomaly generation methods can be well extended to 3D, providing a new idea for subsequent multi - modal data anomaly generation.

\section{Memory Usage and frames-per-second (FPS)} 
\label{ME}
Compared to methods such as 3DSR, CFM. Our BridgeNet balances memory usage and FPS while achieving high accuracy in Table ~\ref{tab_fps}.

\section{Visualization Results}
\label{VR}

We first provide qualitative result images of additional samples from the MVTec-3D AD  dataset in Figure ~\ref{figs1} to further demonstrate that BridgeNet can effectively detect and localize various combinations of anomalies across different modalities. Furthermore, we present qualitative analysis for all sample categories in the Eyecandies dataset in Figure ~\ref{figs2}. To better demonstrate the anomaly localization capabilities of BridgeNet, we selected prediction results from samples with more diverse anomaly locations.

\end{document}